# Artificial Intelligence and Design of Experiments for Assessing Security of Electricity Supply: A Review and Strategic Outlook

*Priesmann, J.[a,*], Münch, J.[b], Ridha, E.[a], Spiegel, T.[b], Reich, M.[b], Adam, M.[b], Nolting, L.[a], Praktiknjo, A.[a,*]*

*[a]Chair for Energy System Economics, E.ON Energy Research Center/School of Business and Economics, RWTH Aachen University, Mathieustr. 10, 52074 Aachen, Germany*

*[b]University of Applied Sciences Duesseldorf, Centre of Innovative Energy Systems, Muensterstr. 156, 40476 Duesseldorf, Germany*

**Corresponding authors: jan.priesmann@eonerc.rwth-aachen.de, apraktiknjo@eonerc.rwth-aachen.de*

## Abstract

Assessing the effects of the energy transition and liberalization of energy markets on resource adequacy is an increasingly important and demanding task. The rising complexity in energy systems requires adequate methods for energy system modeling leading to increased computational requirements. Furthermore, with complexity, uncertainty increases likewise calling for probabilistic assessments and scenario analyses. To adequately and efficiently address these various requirements, new methods from the field of data science are needed to accelerate current methods. With our systematic literature review, we want to close the gap between the three disciplines (1) assessment of security of electricity supply, (2) artificial intelligence, and (3) design of experiments. For this, we conduct a large-scale quantitative review on selected fields of application and methods and make a synthesis that relates the different disciplines to each other. Among other findings, we identify metamodeling of complex security of electricity supply models using AI methods and applications of AI-based methods for forecasts of storage dispatch and (non-)availabilities as promising fields of application that have not sufficiently been covered, yet. We end with deriving a new methodological pipeline for adequately and efficiently addressing the present and upcoming challenges in the assessment of security of electricity supply.

## Nomenclature

### Abbreviations

| | |
|---|---|
| ACER | Agency for the Cooperation of Energy Regulators |
| AI | Artificial intelligence |
| ANN | Artificial neural network |
| API | Application programming interface |
| CNN | Convolutional neural network |
| DOE | Design of experiments |
| ERAA | European resource adequacy assessment |
| FFNN | Feed-forward neural network |
| LDA | Linear Discriminant Analysis |
| LSTM | Long short term memory network |
| MARS | Multivariate adaptive regression splines |
| PCA | Principal Component Analysis |
| PLEF | Pentalateral Energy Forum |
| RAM | Random access memory |
| RES | Renewable energy sources |
| RNN | Recurrent neural network |
| SVM | Support vector machine |
| TSO | Transmission system operator |

## 1 Introduction

Assessing the security of electricity supply is an increasingly important and demanding task. In particular, depicting the effects of the energy transition and liberalization of energy markets on resource adequacy is becoming more relevant and challenging. This is mainly due to the rising complexity in energy systems [1] calling for efficient and adequate methods for energy system modeling (see e.g. [2] for the case of optimization models). As a direct reaction to this, the *European Agency for the Cooperation of Energy Regulators* (ACER) has proposed a comprehensive set of requirements that assessments of resource adequacy in the European context should fulfill: the so-called *Methodology for the European resource adequacy assessment* (ERAA methodology [3]). While high standards for assessments of security of electricity supply are formulated in that document, their implementation in practice comes with some challenges. These include the following:

- conducting prognoses of electricity demand for all countries that are part of the European interconnected grid in future scenarios,
- forecasting unavailabilities of power plant units accounting for *common-mode* events[1] and temporal linkages[2],
- simulating international power flows,
- depicting storage dispatch,
- accounting for climate change in the weather models,
- simulating market mechanisms that cause incentives for (des-)investments in electricity assets, and
- appropriately representing uncertainties in the aforementioned areas.

These challenges can be summed up in three key points: (1) Improving the availability and the quality of input data, (2) improving data forecasts, and (3) reducing the computational complexity of models to allow incorporating the additional requirements listed above into the probabilistic assessment models.

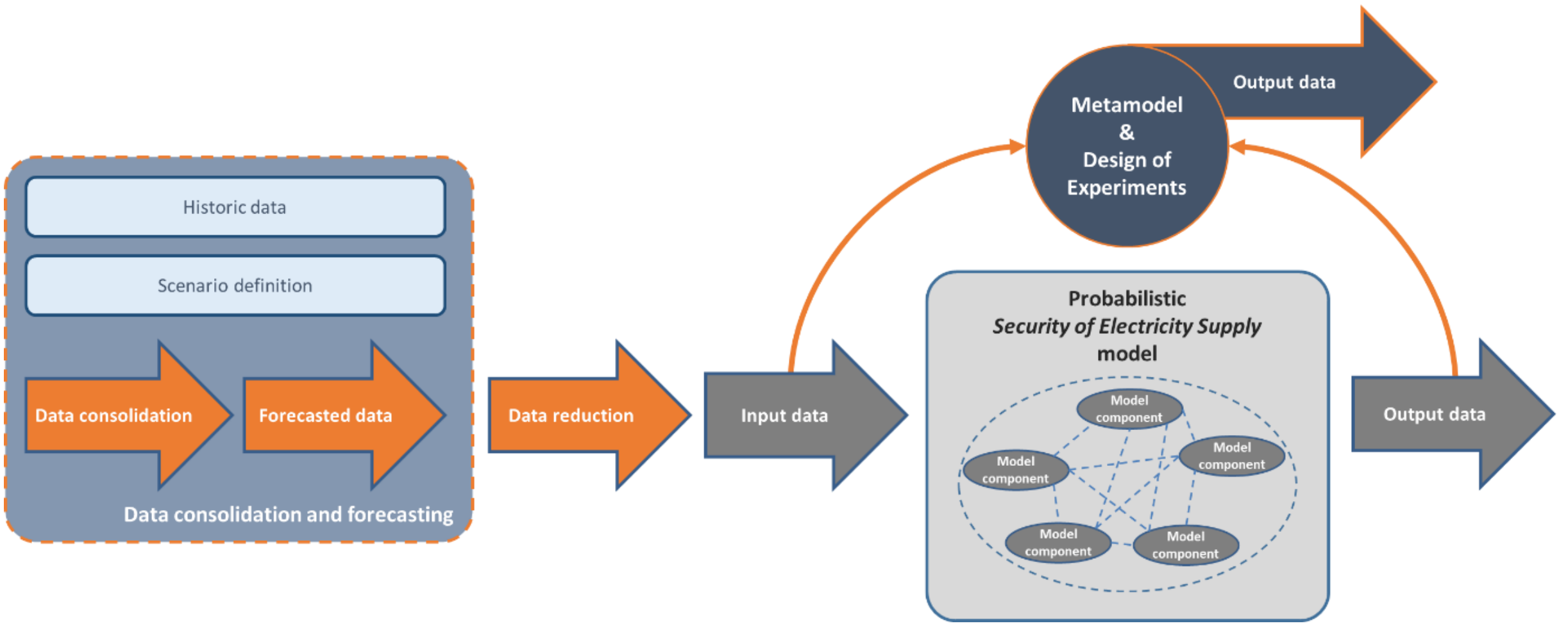

*Figure 1: Pipeline for the assessment of security of electricity supply incorporating metamodeling and design of experiment.*

At this point, there is considerable potential for new methods from the field of data science to address these challenges which is illustrated in Figure 1. Data availability and quality can be improved by, e.g., applying methods for data consolidation [4]. As the assessment of security of electricity supply is usually used for strategic decisions, respective models need to rely on forecasts. Forecasting methods can be applied to a multitude of input data [5,6]. Possible fields of application are the forecasting models for electricity load (see e.g. [7]) or renewable feed-in time series (see e.g. [8]).

[1] These are events that cause the joint unavailability of various assets, like cold spells.
[2] These are events that are caused by preceding failures.

Probabilistic models for the assessment of supply security that tackle the aforementioned challenges turn out to be computationally complex (i.e. requiring a high amount of computational resources such as core-hours or random access memory (RAM)). The detailed analysis of different future scenarios is therefore limited by the necessary hardware availability and computing time. Hence, only a few scenarios can be evaluated adequately. For reducing the model complexity, the reduction of input data [9], the reduction of depicted systemic complexity [10], and metamodeling approaches can be applied. For the case of metamodeling, Nolting et al. [11] showed that particularly approaches from the fields of artificial intelligence (AI) and design of experiments (DOE) seem to be promising for mapping the relationships between model input variables and model results without encountering limitations in terms of available computing resources.

However, the overall potential of AI-based methods in the context of the assessment of security of electricity supply in systems with high shares of renewable energy sources (RES) has not yet been systematically evaluated. Hence, the goals of this review are to (1) identify relevant methods and algorithms from the field of AI and DOI, (2) to associate potential fields of application, and (3) to synthesize the findings and provide a strategic outlook on how to beneficially embed AI-based methods within the assessment of security of electricity supply. By conducting a review of existing approaches and providing an outlook on their potential to enhance resource adequacy assessments, we substantially contribute to the existing body of literature. Figure 2 shows the process applied for conducting the systematic review.

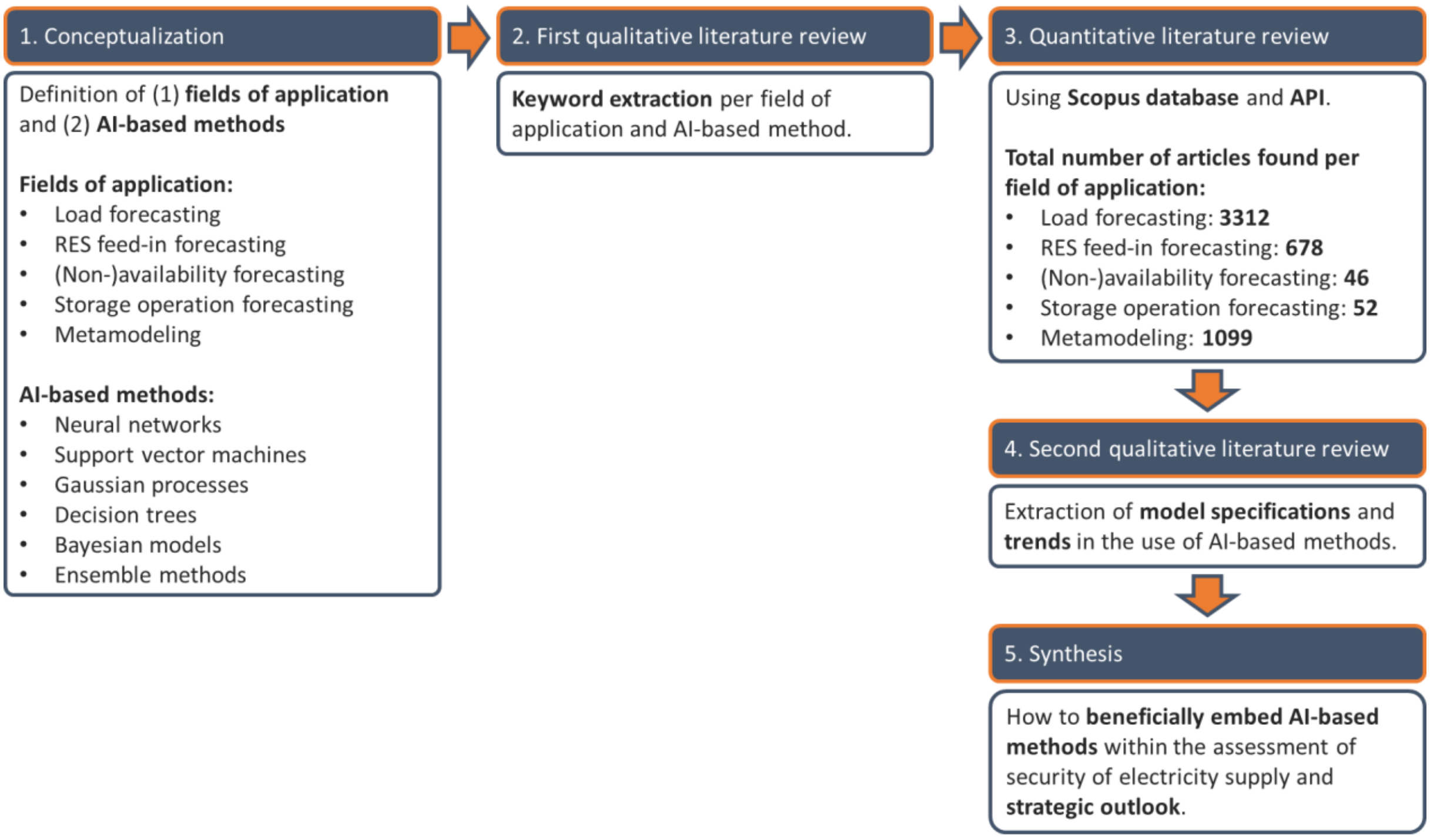

*Figure 2: Overview of systematic review design*

The process of our systematic review is illustrated in Figure 2. We start with the conceptualization of the review by defining and filling the two dimensions (1) fields of application and (2) AI-based methods. These two dimensions are based on previous works such as [11,12]. We then conduct a first qualitative literature review to identify keywords and assign the keywords to the categories defined in the first step. Using these keywords, we conduct a large-scale literature review using the *Scopus* database, the provided application programming interface (API), and the package *pybliometrics* [13]. After generating our article database, we conduct a second qualitative literature review and identify patterns in model specifications and trends in the use of AI-based methods. Finally, we summarize our findings across all fields of application and AI-based methods.

The remainder of this manuscript is structured as follows: In section 2, we provide a brief overview of methods to assess security of electricity supply. In section 3, we review existing AI methods. The metamodeling approach is introduced in section 4. Section 5 is dedicated to identifying fields of application regarding the AI-based forecast. In section 6, we demonstrate metamodeling to be another promising field of application and show its linkage to DOE. We conclude in section 7 by summarizing our findings and deriving their implications for the applicability of AI and DOE for assessments of security of electricity supply.

# 2 Assessing the security of electricity supply

While security of electricity supply covers multiple dimensions and there is a broad range of definitions (see [14], we will focus on resource adequacy in the sense of the *ex ante* evaluation of the energy system's ability to cover electricity load. For assessing this *ex ante* perspective, two general approaches can be distinguished: On the one hand are rather straightforward deterministic capacity balances between secured feed-in power and electricity load during the hour of peak load. On the other hand, complex probabilistic simulations in hourly resolution are used to determine key figures of supply security under consideration of stochastic influences on (1) the availability of fossil power plant blocks, (2) the fluctuating feed-in of renewables, and (3) electricity load. Both approaches have been applied in various studies by consulting companies, research institutions, and transmission system operators (TSOs) in different contexts. Table 1 provides an overview of existing studies, methods used and core results achieved[3].

*Table 1*: Summary of literature review on recent studies in the field of assessing the security of electricity supply

| Reference | Methodology | Geographical scope | Time horizon | Key findings |
|---|---|---|---|---|
| [15] | Deterministic capacity balance | Germany and neighboring countries | 2020, 2030 | Focused capacity markets are necessary to ensure security of supply |
| [16] | Probabilistic simulation | Germany and neighboring countries | 2013-2035 | Energy Only Markets can guarantee security of supply in central Europe |
| [17] | Deterministic capacity balance | European Union | 2016, 2020, 2025 | High share of RES increases pressure on security of supply |
| [18] | Deterministic simulation (Power2Sim model) | European Union | 2023 | It is not necessary to maintain power plant capacities as a reserve |
| [19] | Probabilistic simulation | Germany and neighboring countries | 2015, 2025 | Security of electricity supply is at high levels in future scenarios for 2025 |
| [20] | Probabilistic simulation | Central Europe (PLEF* region) and neighboring countries | 2009-2014 (ex-post), 2030 (ex-ante) | International dependency of security of supply increases |
| [21] | Probabilistic simulation | Germany and neighboring countries | 2020, 2023, 2025 | Supply shortages in Northern Germany from 2023 at the latest, in Southern Germany from 2025 at the latest |
| [22] | Probabilistic simulation (Monte-Carlo) | European Union | 2020, 2025 | International dependency of supply security, supply shortages are expected in Germany in 2025 |
| [23] | Deterministic capacity balance | Germany | 2020, 2023 | Security of supply in Germany is not affected if the 20 oldest lignite-fired power plants are shut down |
| [24] | Deterministic simulation | European Union | 2018/19 and 2020/21 | There is a need for grid reserve beyond 2020 |

[3] As can be seen, the regional focus is set on Europe, here.

| | | | | |
|---|---|---|---|---|
| [25] | Probabilistic simulation | Central Europe (PLEF region) | Winter 2018/19 and winter 2023/24 | In winter 2023/24 the security of supply in Germany, Luxembourg, and the Netherlands is at risk |
| [26] | Probabilistic simulation | Germany (divided into North and South) and neighboring countries | 2025 | Electricity supply in Germany in 2025 is secured, but Southern Germany will depend on imports |
| [27] | Deterministic capacity balance | Germany and European Reserves | 2017, 2020, 2023 | Significantly accelerated coal phase-out in Germany is possible without endangering the security of supply |
| [28] | Probabilistic simulation (Monte-Carlo) | European Union | 2025 | No supply shortages are expected in most European countries; only Finland, Greece, and Ireland will face problems |
| [29] | Deterministic capacity balance | Germany | 2017-2021 | In 2021, Germany depends on electricity imports during peak load. Before 2021, national capacities are sufficient. |
| [30] | Probabilistic simulation (Monte-Carlo) | European Union | 2020, 2025 | Supply interruptions are to be expected in central Europe in capacity reduction scenarios |
| [31] | Probabilistic simulation (Monte-Carlo) | European Union | 2021, 2025 | Loss of load to be expected in countries with high penetration of coal-fired power plants in a *low carbon sensitivity scenario.* |
| [12] | Probabilistic simulation | Germany and neighboring countries | 2020, 2022, 2023 | Turning away from the absolute level of security of electricity supply in Germany. |
| [32] | | European synchronous grid area | 2025, 2030 | In a *baseline scenario* loss of load is only to be expected in Malta, Sardinia, Iceland und Turkey. COVID- 19 has only minor impact on resource adequacy. |

** PLEF = Pentalateral Energy Forum*

Figure 3 summarizes the essential characteristics and common implementations of the two model classes. Here, it can be seen that deterministic capacity balances represent rather straightforward, top-down models to derive non-probabilistic key figures such as capacity margins. They are usually conducted for one hour per year (i.e., the hour with the highest electricity load) and consider only one (often worst-case) weather situation. On the other hand, probabilistic simulation models represent rather complex, bottom-up models that are used to calculate stochastic key figures such as expected loss of load durations per year. They are commonly performed in hourly resolution and reflect different weather situations (so-called historic weather years[4]).

[4] A weather year represents the meteorological conditions in an area and is used to calculate weather dependent electricity load and feed-in profiles. See e.g. [7].

**Deterministic capacity balance**

- **Top-down approach**, i.e. modeling is based on a high level of emergence
- **Low model complexity**, i.e. effort and costs for implementation and computing time are virtually negligible.
- Calculated **key figures of supply security:**
  - ***Remaining Capacity (RC)***: Difference between the available capacity and peak electricity load. This surplus is available to cover unexpected loads and to compensate for power plant outages that have not been accounted for.
- **Common approach**:
  - Focus on **peak load hour**
  - Consideration of one **(worst-case) weather year**

**Probabilistic simulation**

- **Bottom-up approach**, i.e. modeling is based on sub-elements of the overall system.
- **High model complexity**, i.e. substantial efforts and costs for implementation as well as long computing times and high memory needs should be accounted for.
- Calculated **key figures of supply security:**
  - ***Loss of Load Probability (LoLP)***: Probability of load shortfall during the examined hour.
  - ***Loss of Load Expectation (LoLE)***: Expectation of supply shortages in hours during the scenario year under consideration.
  - ***Expected Energy not Served (EEnS)***: Amount of energy demand in MWh/a that is expected not to be covered during the scenario year under consideration.
- **Common approach**:
  - Consideration of **8,760 hours** per scenario year.
  - Analysis of **different weather years**.

*Figure 3:* Summarizing comparison of modeling approaches. For references, please see Table 1.

From the sheer number of studies, the range of different and often opposing key findings, and the heterogeneity of the authors and principals, it can be concluded that there is a considerable need to provide the scientific basis for sound assessments of security of electricity supply. Further, the band of uncertainty of the results that comes with the different input data calls for assessing a larger variety of scenarios to depict possible future developments. We hence transfer methods from the fields of AI and DOI to contribute to more sound assessments of security of electricity supply.

# 3 Artificial intelligence for energy system analysis

Although there is no commonly agreed-upon definition of the term “Artificial intelligence”, it is typically used to describe behavior exhibited by computers that was initially thought to require (human) intelligence [33]. There is, however, a consensus on the distinction between strong or general AI, which mirrors the capabilities of intelligence as a whole on the one side and weak or narrow AI, which is developed in order to solve specific problems on the other side [34].

The methods explained in this paper are problem-specific approaches that fit under the term of narrow AI. Moreover, they are also part of the realm of machine learning, a subfield of AI concerned with learning statistic relationships. Machine learning can thus also be regarded as a subfield of statistics. Contrary to other statistical measures, the exact nature of the statistic relationship (e.g., a functional relationship) between the input data and the output data is not explicitly defined but rather implicitly inferred by the machine learning model itself.

Data is typically present as a number of samples or observations which have values in certain features [35]. If the data is imagined as a table, the columns typically denote the features, while the rows are the observations. Often, the term “feature space” is used when talking about data in a machine learning context. This stems from regarding $N$ features as the axis of an $N$-dimensional space where every observation is one point in the space. Table 2 gives an example of features and observations for time series: The columns contain the electricity generated through various conversion technologies in Germany on 1st June 2021, with each column corresponding to one feature. The observations are denoted through the time axis.

*Table 2: Realized generation of various conversion technologies during 2021-06-01 in Germany in MWh. Data source: Bundesnetzagentur 2021*

| Time | Biomass | Water | Wind Offshore | Wind Onshore | Photovoltaics |
|---|---|---|---|---|---|
| **00:00** | 1,178 | 528 | 274 | 1,570 | 0 |
| **00:15** | 1,176 | 500 | 277 | 1,561 | 0 |
| **00:30** | 1,165 | 506 | 262 | 1,534 | 0 |
| **00:45** | 1,174 | 499 | 264 | 1,495 | 0 |

Three forms of machine learning can be distinguished by how the model receives feedback on its learning process: Supervised learning, unsupervised learning, and reinforcement learning [36].

In supervised learning, the data itself contains a set of variables to be explained, the so-called *labels*, and data that is used as explanatory input, the so-called *features*. The input data and labels are fed into the model which then learns the relationship between the two [37]. This is referred to as training. A part of the data is usually withheld from the model during training in order to test whether it generalizes well, i.e., whether it correctly predicts labels for data it has not been trained on. It is then possible to predict labels for input data for which no label is available (e.g., future values of a time series). Supervised learning can be further divided into regression, where the label is a metric variable, and classification, where the label is one of several discrete classes [36].

In unsupervised learning, no labels exist and, therefore, problem-specific measures are used to evaluate the model's quality. For example, clustering is the unsupervised counterpart to classification: In clustering, no classes are known *a priori* and the model creates its own classes, which are called clusters.

Finally, reinforcement learning works with a cost function that defines rewards and penalties. This approach is particularly useful when it is impossible to cover all possible system states during training, such as when teaching machines to play games or in autonomous driving. [37]

With regard to energy system modeling, AI methods are used broadly in three fashions: Preprocessing of relevant input data, forecasting of time series, and metamodeling energy system models. In the first case, an AI method (or model) prepares data for other models while in the second case, an AI method is used to forecast relevant data that might be fed into further energy system models or directly analyzed. The third case refers to using a conventional energy system model in order to generate data for training the AI model. That is, the AI model is used to model the conventional model's behavior and to allow for a broader scope of scenarios to be investigated (see [11].

The following section 3.1 provides an overview of AI-based methods used for forecasting, while section 3.2 gives an overview of a selection of AI-based methods for data reduction in the context of energy system modeling. Both subsections are not to be understood as comprehensive reviews of all available AI-based methods, but rather as guides on models and methods that are of particular interest to energy system modelers looking to integrate AI into their research. In section 3.3 methods for evaluating the accuracy of AI-based models are presented. Detailed applications of forecasting methods and AI-based metamodels will be discussed in sections 4 and 5.

## 3.1 Supervised learning methods for data consolidation, forecasting input data, and metamodeling approaches

The methods presented in the following belong to the field of supervised learning, i.e. they are concerned with learning the relationship between input features and a label to be predicted.

The methods can be applied for data consolidation, forecasting, and metamodeling. Data consolidation comprises the handling of data gaps or multiple data sources that need to be merged. Forecasting refers to extrapolating historical information (e.g. on electricity load or renewable feed-in) into the future. Metamodeling simulates the behavior of system models by learning the relationship between the model input and output data (for more information on metamodeling see sections 4 and 6).

### 3.1.1 Artificial Neural Networks

Artificial neural networks (ANNs) are a versatile and powerful tool for forecasting, which are used for a multitude of applications, including image recognition, natural language processing, and time-series forecasting.

The mathematical concept of ANNs is inspired by the human brain: Like a biological brain, a neural network consists of neurons that exchange information [38]. Like a biological neuron, an artificial neuron receives inputs from other neurons, which are weighted and summed. The resulting sum is then put through an activation function (e.g., a sigmoid function). The output of the activation function constitutes the neuron's output, which is in turn part of the next neuron's inputs [39]. Figure 4 shows the structure of a basic neural network.

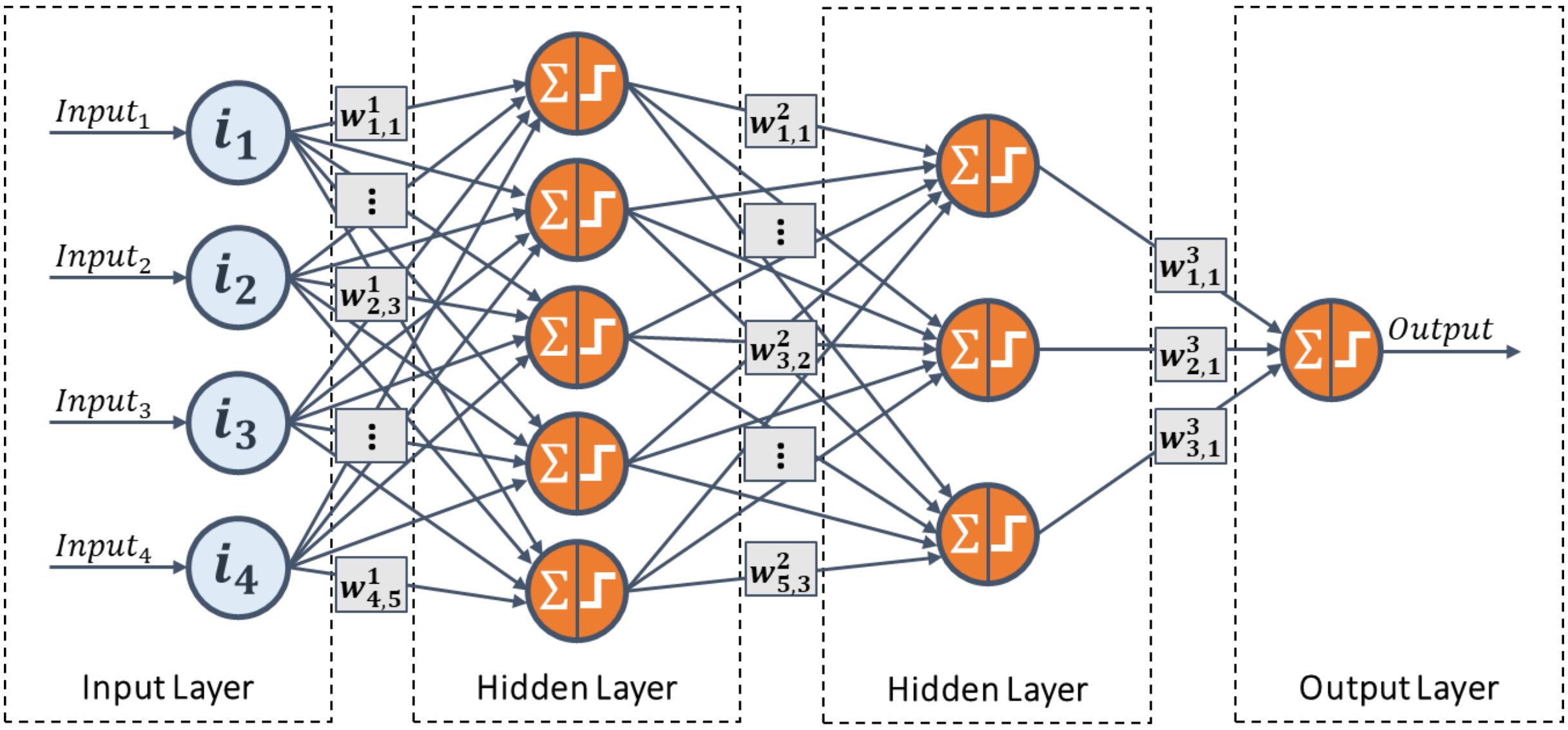

*Figure 4: A basic feedforward neural network consisting of several layers of connected neurons. Here, connections between two neurons have a weight that is determined during training and neurons in the hidden and output layers are equipped with an activation function.*

An ANN consists of several layers, each layer, in turn, comprising one or multiple neurons [40]. All ANNs have an input layer, which receives the inputs fed into the model, and an output layer, which contains the model's output. In addition, more layers may be added between the input and the output layer. These layers are called hidden layers and a model including hidden layers is called a deep learning model [39]. In a so-called densely connected ANN, all neurons within one layer are connected to all neurons in the preceding and the next layer. These networks constitute the basic form of neural networks and are called feed-forward neural networks (FFNNs) or multilayer perceptrons [38]. Other network architectures, which are particularly well-suited for specific purposes, have been developed and will be briefly described next.

**Convolutional Neural Networks:** Convolutional Neural Networks (CNNs) are a type of neural network particularly suited for dealing with data that is inherently structured in a grid-like fashion [41]. In general application, CNNs are most known for having been used with great success in image recognition and image classification [40]. In energy system modeling, they find use for example as tools for detecting and classifying power quality disturbances (e.g.

[42,43]), in order to prepare data for other models (e.g. [44,45]), or for feed-in and load forecasting (e.g. [42,46]).

CNNs make use of convolution, a mathematical operation whereby one mathematical function is averaged using a second function. In machine learning, convolution is not performed on continuous functions, but discrete data, making it a matrix multiplication [38]. During convolution, the input data is passed over with a kernel, which can be understood as a set of weights with which every entry in the input data and its surrounding entries are weighted and then summed. After the convolution, an activation function (as in fully connected, linear ANNs) is employed and the data is pooled. Pooling maps the input data to a reduced output. Maxpooling, for example, outputs the maximum of the values held in a set of neurons. Pooling helps to make the network less sensitive to small variations [40]. The combination of convolution and pooling in CNNs allows recognizing more and more abstract features in later layers of the network [43].

**Recurrent neural networks:** Recurrent neural networks (RNNs) are particularly suited for dealing with sequence-like data, such as time series, written or spoken language. This has led to this network architecture being used in a variety of energy-related time series applications, such as forecasting weather data (e.g. [47]), renewable feed-in (e.g. [48]), or electricity load (e.g. [49–51]).

RNNs are named after the recurrent connections employed in them, which feed a neuron's output from prior steps back into itself as new input [52]. This allows long-term dependencies to form [39]. As a result, as opposed to plain feedforward networks, RNNs can deal with sequences of arbitrary lengths.

Long short term memory networks (LSTMs) are a subtype of RNNs that allow links between time steps other than $t$ and $t-1$ [53]. There are several ways to achieve this, with LSTMs allowing the model to learn when to include information from prior time steps rather than determining this manually [38]. The neuron is able to accumulate knowledge over time and also to "forget" it over time by selectively allowing prior outputs to influence the current computation. This approach has proven particularly effective in a variety of areas concerned with time series prediction, including speech recognition, language translation, and image captioning [38,40].

### 3.1.2 Support Vector Machines

Support vector machines (SVM) are a method mainly used for binary classification. The principle derives from the idea of creating a hyperplane that divides the feature space into two areas, each of which contains the observations belonging to one class [54]. The hyperplane in an $n$-dimensional space is always $n-1$-dimensional. For example, in a two-dimensional space, the hyperplane is a line dividing the space into two areas, with each class lying on one side of the line [55]. Support vector machines are an extension of the hyperplane approach that remedies some of its drawbacks, such as the inability of a linear hyperplane to correctly depict a non-linear division of observations [36,37,55]. Although originally intended for classification tasks, extensions of SVMs exist that allow dealing with regression tasks as well, including non-linear regression [56]. This makes them a flexible tool that finds use in a variety of energy-related forecasting contexts, for example in predicting renewables' feed-in (e.g., [57]), building energy consumption (see [58]) or grid investments (e.g., [59]).

### 3.1.3 Gaussian process regression

The Gaussian process is a stochastic process in which every random variable is assumed to be multivariate normal distributed. As a multivariate normal distribution is defined by a mean vector and a covariance matrix, a Gaussian process is defined by a mean and a covariance function. Generally speaking, the covariance function describes the similarity between the random variables and defines the smoothness of the function [60]. When a Gaussian process regression (GPR) is applied to supervised learning problems, it provides a distribution over functions, inherently utilizing uncertainties. GPR can be used for a variety of tasks in a supervised learning setting, for example, anomaly detection [61] or as a basis for model predictive control [62]. Due to computational limitations, GPRs are generally used on small to medium-sized data sets, but recent work is exploring ways into handling big data problems [63].

### 3.1.4 Transformers

Another deep learning methodology is the so-called transformer (a.k.a. X-former), which was presented at the Neural Information Processing Systems conference in 2017 (Vaswani *et al.*, 2017). Transformers are mainly used in natural language processing, computer vision, speech processing, and audio processing and using the mechanism of "attention". In this area, for example in aspect-level sentiment classification, attention mechanisms have also been added prior to LSTMs (AT-LSTMs) to achieve better results (Wang *et al.*, 2016). For various use cases, Transformer-based pre-trained models (Qiu *et al.*, 2020) can achieve state of the art, so they are preferred especially in the field of natural language processing. (Lin *et al.*, 2021)

The transfer of the methodology to the forecasting of time series is discussed [64]. Here, transformers are used to predict synthetic and real-world datasets (electricity and traffic). In this work, two weaknesses of transformers were uncovered when used to predict time series. Direct modeling of long time series is not feasible because the space complexity of the canonical transformer grows quadratically with sequence length (leading to a memory bottleneck). Second, there is a susceptibility to anomalies in time series due to the insensitivity of the pointwise dot-product self-observation in the canonical transformer architecture to the local context. This local context provides information on whether the pattern of the time series is changing due to an event (e.g., holiday), a change point, or an anomaly. The authors present two potential solutions to these problems. It is proposed that problem of susceptibility to anomalies in time series can be solved using convolution self-attention. To remove the memory bottleneck, LogSparse transformers are proposed, which reduce the dot products to be calculated. Based on their results, Li *et al.* (2019) conclude that transformers can capture long-term dependencies better than LSTMs.

### 3.1.5 Decision Trees

Decision trees are a tool used in regression and classification that represents a sequential perspective on machine learning [65]. The learning process of a decision tree works by repeatedly dividing the data depending on the input features and assigning a label to the groups of observations created in this way. Starting from the whole feature space, one of the features is selected and a threshold value for this feature is defined, dividing the data into two groups. Each observation is then assigned to one group, depending on which side of the threshold it lies on [66]. This way, the feature space is divided into two subspaces, each of which is then assigned a predicted label (e.g., the mean label of its observations in regression or the most frequent class of its observations in classification). The feature and threshold which serve as a decision boundary are selected according to which selection minimizes the prediction error [55].

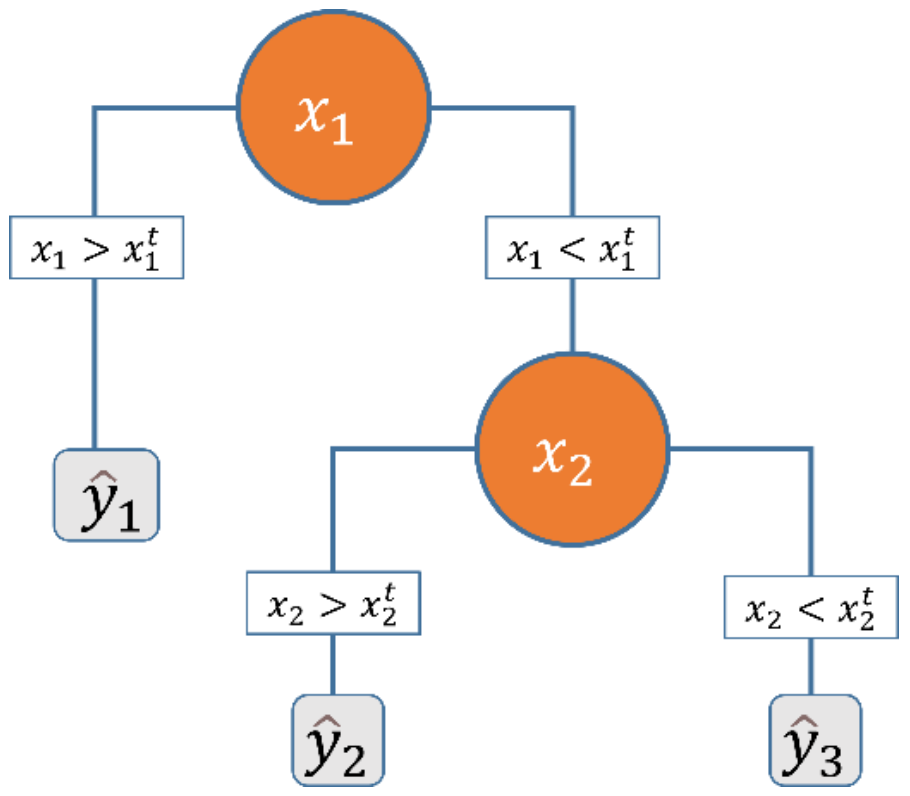

*Figure 5: Exemplary decision tree structure: The label $y$ is to be predicted using the input features $x_1$ and $x_2$. In the first step, the decision tree partitions the observations according to whether its value in $x_1$ exceeds the threshold $x_1^t$. If this is the case, the observations are assigned the predicted label $\hat{y}_1$.Otherwise, there is a further decision taken: The other observations are further divided according to whether their values in $x_2$ exceed the threshold $x_2^t$ and are assigned the predicted labels $\hat{y}_2$ and $\hat{y}_3$, respectively.*

This process of feature space division is repeated iteratively, leading to smaller and smaller subgroups with finer predictions [55]. This can be visualized in a hierarchical tree structure, giving decision trees their name. An exemplary visualization of a decision tree is given in Figure 5. Their capability of being easily visualized makes decision trees easily interpretable, which is one of their greatest strengths [66]. Like other supervised learning methods, in energy system analysis, they find use in various forecasting contexts, e.g. in predicting buildings' energy consumption (e.g., [67]).

### 3.1.6 Ensemble Methods

Ensemble methods are hybrid solution strategies. The idea of hybridization is based on the no-free-lunch theorem [68]. The theorem states that there is no single optimal algorithm for every optimization problem. A hybridization strategy utilizes multiple algorithms and combinations of model results to improve optimization techniques [69] aiming for better overall model performance in terms of the speed-accuracy-complexity tradeoff [70]. The ensemble training process can be distinguished into bootstrap aggregation (bagging) and boosting strategies [69] which are described in the following.

**Bootstrap aggregation (Bagging):** Bootstrapping describes a resampling method to train and validate models by using random subsets of the data set [71]. Bootstrap aggregation (bagging) is a method that (1) trains multiple models in parallel, i.e. independently based on different bootstrap samples (data subsets), and (2) creates an overall prediction by averaging all model prediction results [72]. Averaging predictions of several bootstrap sample models reduces the variance component of the overall generalization error [66,73].

The bagging method can be applied to various model architectures, e.g. SVMs (Kim *et al.*, 2003; Drucker and Cortes, 1996), k-nearest neighbors [74], or random forests [75]. Random forest models have become the most popular approach for applying bagging to decision trees [55,73]. This is based on the extension of the bagging process in terms of resampling the data by its samples and features.

**Boosting:** Boosting is an ensemble method to train multiple models sequentially. Successive models attempt to optimize the overall model performance based on the knowledge of previous model error [76–78]. This approach is different from bagging where models are trained in parallel without knowledge of the performance of other trained models.

The boosting ensemble learning method is commonly used in different variations, i.e. AdaBoost, Gradient boosting machines, stochastic gradient boosting. AdaBoost trains a series of models adaptively aiming to minimize residuals, i.e. previous model error [79]. Gradient boosting machines generalize AdaBoost by minimizing an arbitrary differentiable loss function, so-called pseudo-residual, as objective utilizing the gradient descent method [80]. Furthermore, the stochastic gradient boosting approach expands the gradient boosting machines and incorporates the resampling of samples and features at each training iteration [81].

The boosting method can be applied to various model architectures, e.g. decision trees [78,80–82], neural networks [79,83,84], or naïve Bayes [85].

## 3.2 Reducing the size of model input data using AI-based methods

Preprocessing data has a wide variety of purposes, such as consolidating multiple data sources, filling data gaps, reducing computation time by aggregating data fed into an energy system model, or boosting a subsequently used machine learning model's prediction quality by filtering and preparing relevant features.

### 3.2.1 Time series decomposition and feature selection

Energy system modeling usually involves handling time series data. Time series are in turn often the result of several effects superimposed over each other. Thus, decomposing a time series into its components (e.g. constant, cyclical, and trend components) can help interpret and model time series data [86]. Recently, there has been research conducted focusing on the combination of decomposition and feature selection as preparation for a forecasting model (see e.g. [87] and [88]).

Figure 6 shows an exemplary seasonal decomposition of an electricity load time series. The original time series (orange line) is decomposed into three components: a trend component, a seasonal component, and a residual. In this example, multiplicative decomposition is performed. The multiplication of all components results in the original time series.

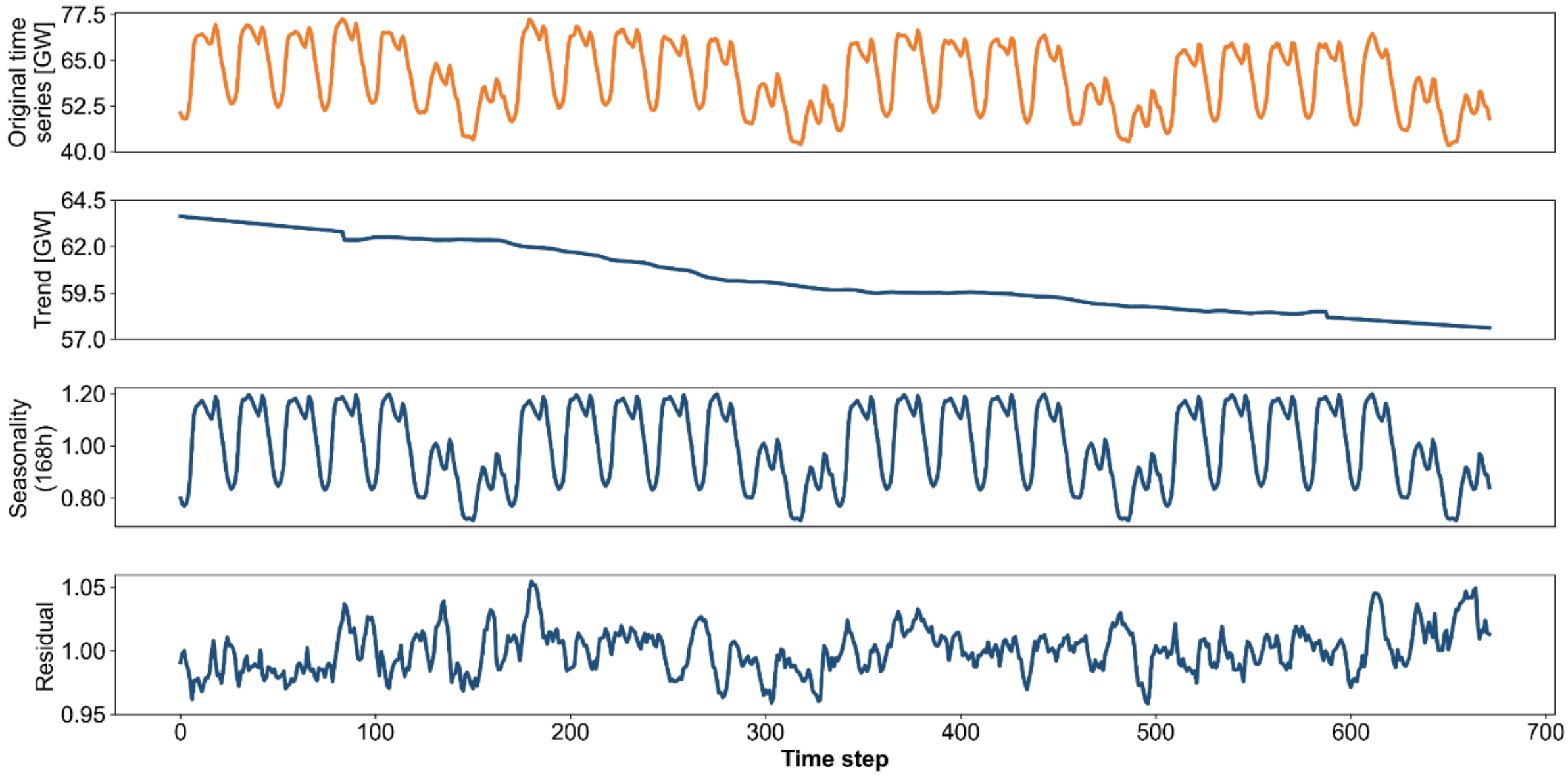

*Figure 6: Trend, seasonality (168 hours), and residual series of a decomposition of a four-week electricity load time series (orange line).*

Feature selection refers to deciding on which features to feed into the model [89]. This is particularly potent in combination with time series decomposition methods, as it can allow keeping only those components of a time series that are both meaningful and predictable. This can increase the quality of the prediction while not needlessly adding unnecessary dimensions to the input data. Typically, feature selection approaches are classified as filters, wrappers, or embedded methods [89]. A filter is independent of the subsequent model the selected features are fed into, while a wrapper "wraps around" a predictive algorithm [90]. The inner algorithm's performance when being fed different selected features is used to decide which features are selected. Embedded methods are predictive methods that inherently include some sort of learning that can be used for feature selection [35].

### 3.2.2 Clustering

Clustering is useful in energy system modeling as a tool for reducing the amount of data fed into the energy system model. Clustering unveils the structure inherent in the data by allocating the observations to distinct groups. Generally, the aim is to find clusters of observations that are as dissimilar to each other as possible while the objects (or observations) within the clusters are as similar to each other as possible [91]. Usually, these similarities are measured by a (dis)similarity or distance measure [92]. The clusters are not known beforehand but rather formed by the algorithm during the clustering process. Thus, clustering is an unsupervised learning technique: there is no *a-priori* known correct cluster allocation.

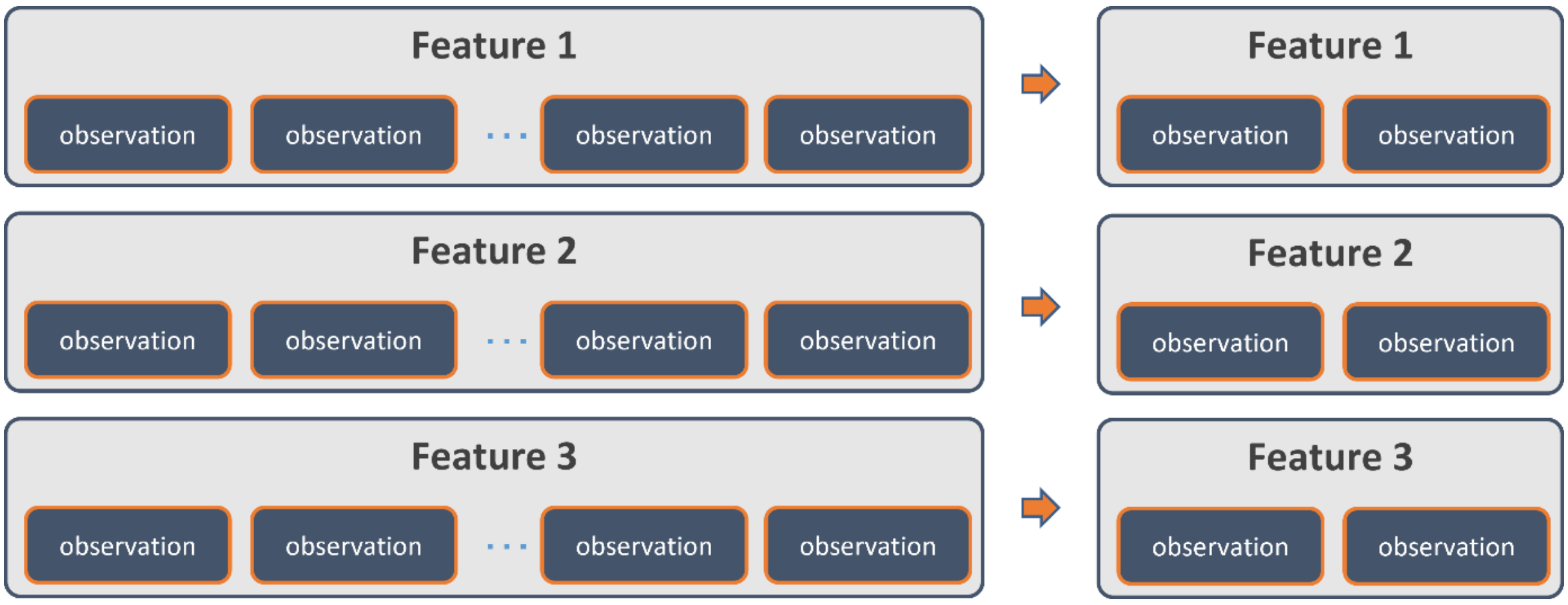

*Figure 7: Schematic of clustering: While the number of features remains identical, the number of observations in the features is reduced.*

Clustering data has a variety of uses: On the one hand, it can help identify patterns that are not easily discernable in the original dataset. On the other hand, clusters can reduce the amount of data by identifying representative cluster centers (typically expressed through the mean, centroid, or medoid of the observation's features). In energy system analysis, it often finds use as a data reduction method, particularly for time series data. Clustering time series (i.e. time series aggregation) allows representing the data through a considerably lower number of data points, which can reduce model runtimes while also impacting model accuracy. This principle is visualized in Figure 7. Hoffmann et al. [9] present a comprehensive review of time-series aggregation methods and compare various approaches, including clustering. Other reviews include the ones conducted by Teichgraeber et al. [93] and Kotzur et al. [94]. Moreover, clustering is used to find representative generation units (e.g. [95]) and customer groups (e.g. [96]).

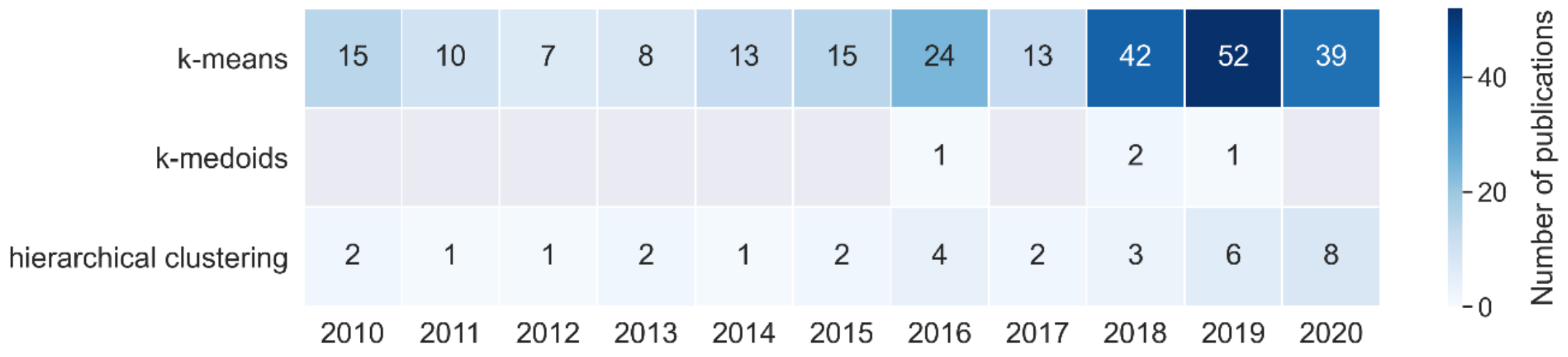

*Figure 8: Number of scientific articles found on the application of popular clustering algorithms in energy system analysis published in energy-related journals in the Scopus database from 2010 to 2020.*

Figure 8 shows the number of articles on some of the most popular clustering algorithms in the field of energy system modeling since 2010[5]. A variety of clustering algorithms exist and choosing a well-suited algorithm is a highly data- and problem-specific task. Therefore, we focus on a brief introduction of two of the most popular clustering approaches: k-means and hierarchical clustering. There is a considerable body of literature on clustering (e.g. [91], [92], or [97]), which the reader may consult for a more comprehensive overview.

**k-means:** k-means is a widely used clustering approach that creates $k$ clusters by allocating observations to cluster centers in such a fashion that the overall sum of distances between the data points and their nearest cluster centers is minimized [36]. In a first step, $k$ cluster centers are initialized. Then, all observations are allocated to the cluster center nearest to them (measured in Euclidean distance), forming the initial clusters. Next, the algorithm recalculates the cluster centers' coordinates, i.e. the "column means" across all observations allocated to each cluster. This process of allocating observations to clusters and recalculating the cluster centers is repeated iteratively until a convergence criterion is met [91].

The advantage of k-means is its ease of use combined with its tendency to create evenly sized clusters. However, not all data is suited to be evaluated using *Euclidean distance*. In particular, non-metric data is difficult to cluster with k-means. In addition, the desired number of clusters $k$ needs to be specified in advance, which demands a priori knowledge on the dataset [91]. Some heuristics exist in order to determine suitable cluster numbers.

k-medoids is a clustering approach similar to k-means. The main difference between the two approaches are: (1) in k-medoids, the representative cluster centers are not calculated as means, but rather, selected out of the elements that form the cluster and (2) k-medoids is compatible with similarity measures other than *Euclidean distance* [98]. However, k-means is still the vastly more popular clustering algorithm in energy system analysis, as evidenced by Figure 8.

**Hierarchical clustering:** Hierarchical clustering gains its name due to the fact that it results in a hierarchy of clusters. There are two approaches towards this: Agglomerative and divisive clustering. In agglomerative clustering, the algorithm starts by assigning each observation to its own cluster. Next, the similarities between all clusters are calculated and those two clusters that are the most similar to each other are merged. The merging of clusters is repeated until a desired number of clusters is reached. Similarly, in divisive clustering, all observations are initially part of one cluster, which is subsequently divided into subclusters [91].

There are several ways of determining which clusters are most similar [91,92]. This makes hierarchical algorithms flexible but requires an in-depth understanding of the data. Therefore, a certain level of expertise and heuristics are needed to cluster data well with hierarchical algorithms.

[5] For details on the search term specification, see Table 10 in the appendix.

In energy system modeling, hierarchical clustering is often used for time series aggregation. Liu and Sioshansi [99], for example, employ hierarchical clustering approaches in order to find representative time periods for capacity modeling. Hoffman et al.'s [9] review also includes time series aggregation methods based on hierarchical clustering.

### 3.2.3 Dimensionality reduction

While data in a two-dimensional space (i.e., with two features) can be easily visualized and understood, data with tens of dimensions or more is impossible to understand visually. In addition, high-dimensional data often poses problems (curse of dimensionality, see e.g. [100]), is difficult to handle mathematically, or carries with it a high computational cost [101,102]. For example, distance measures like those used for clustering lose meaning in high-dimensional spaces since the data becomes naturally sparse [103]. That is why finding a representation of high-dimensional data in a lower-dimensional space is often desirable. This is called reducing the dimensionality of the feature space. Thus, if the reduced-space representation is used for further algorithms and models (such as an ANN), the data to handle is reduced, resulting in faster computation times. Figure 9 contains a visual representation of this principle.

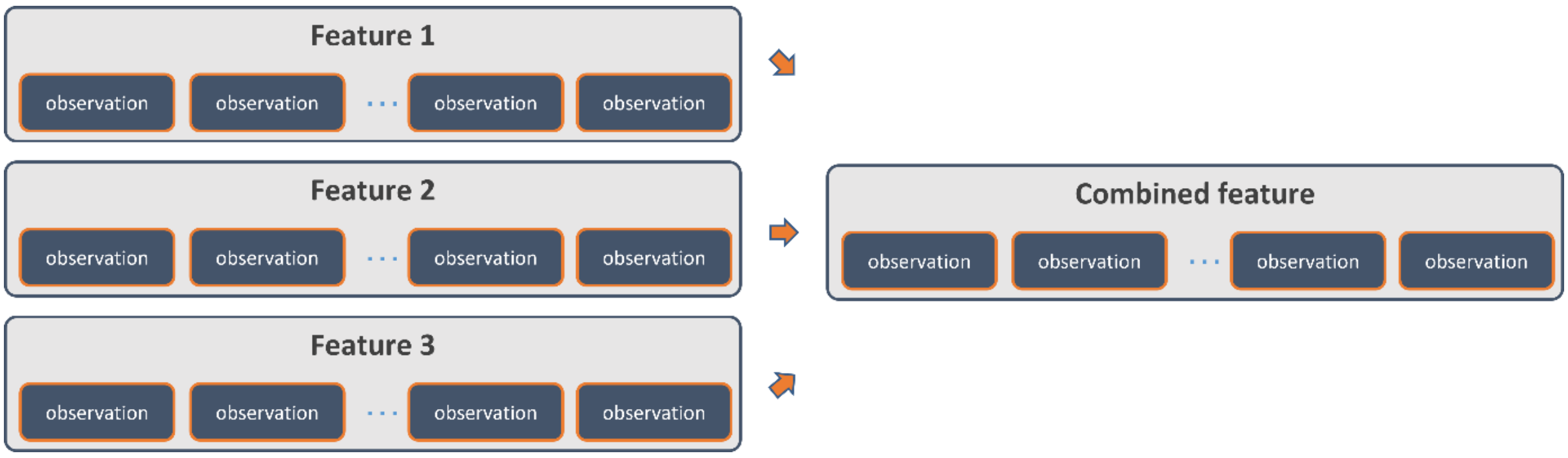

*Figure 9: Schematic of dimensionality reduction: The reduction of data is achieved through reducing the number of features, while the number of observations remains identical.*

There are several tools available to achieve this, widely-spread examples of which are Principal Component Analysis (PCA) and Linear Discriminant Analysis (LDA). Figure 10 shows the number of articles on some of the most popular methods for dimensionality reduction in the field of energy system analysis since 2010[6].

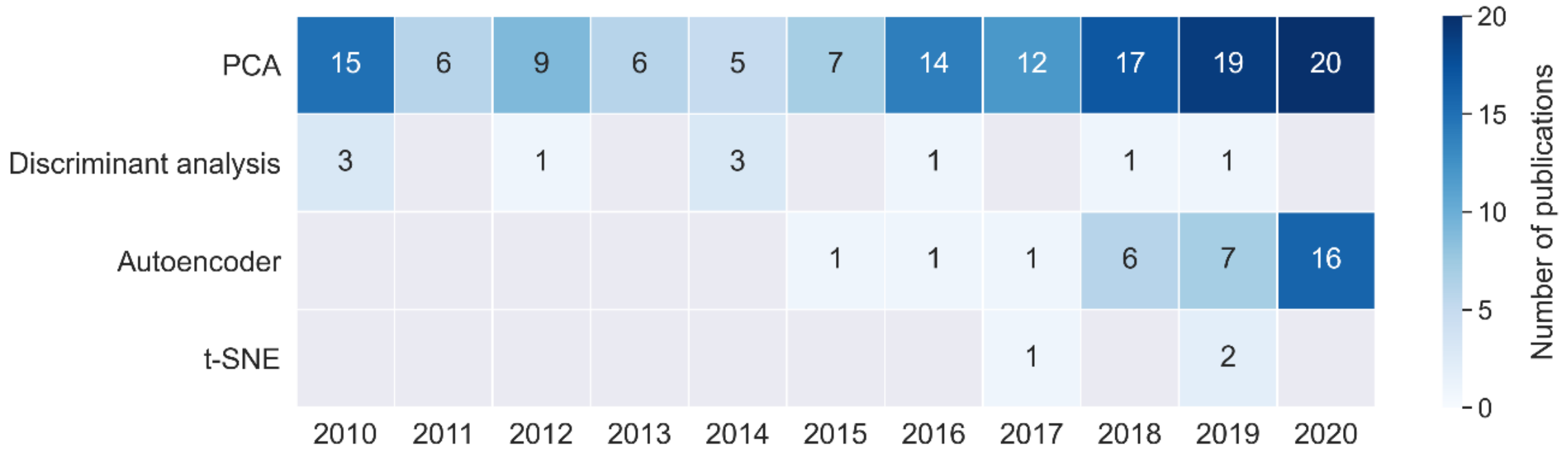

*Figure 10: Number of scientific articles found on the application of popular dimensionality reduction methods in energy system analysis and published in energy-related journals in the Scopus database from 2010 to 2020.*

[6] For details on the search term specification, see Table 11 in the appendix.

**Principal Component Analysis:** PCA assumes that the data's variance determines its meaningfulness. Thus, the lower-dimensional representation of the data is found by maximizing the data's variance [101]. This is done by identifying those "directions" in the data that possess the most variance, the so-called principal components.

The principal components are created as linear combinations of existing features [55]. Mathematically, this is done by performing an eigenvalue decomposition on the data's covariance matrix [101] or by performing singular value decomposition (SVD) to the data itself, which tends to yield more robust results [104]. The user can select the number of dimensions they wish to transform the data into.

**Linear Discriminant Analysis:** LDA is a classification method mapping metric input variables to a discrete output variable. It works in a two-step fashion: The data is first transformed and then classified. The transformation step is similar to PCA in that a new space is created, the axes of which are linear combinations of the original features [105].

Next to conventional methods of dimensionality reduction, other, novel tools have recently gained popularity. Two of those are t-Distributed Stochastic Neighbor Embedding (t-SNE) and Autoencoders

**t-Distributed Stochastic Neighbor Embedding:** T-SNE was created by can der Maaten et al. [106], intended as a tool for visualizing data. Nonetheless, it can also be used as a dimensionality reduction tool. For example, [107] have used t-SNE in the course of forecasting wind feed-in.

**Autoencoders:** An autoencoder is a type of deep neural network design that has been existing for several decades [38,108–112]. The design has been seeing renewed interest throughout recent years due to its various application areas, e.g. dimensionality reduction, feature extraction or anomaly detection, and its capability of preserving and representing non-linear relationships of input data.

The fundamental structure of an autoencoder is based on (1) an encoder to map the input space with a linear or non-linear transformation to a lower-dimensional latent space and (2) a decoder to map the salient input features represented by the latent space back to reconstruct the input space.

Formally, an input example $\mathrm{x} \in \mathrm{X}$ with $\mathrm{X} \in \mathbb{R}^n$ is mapped to a hidden representation $h(x) \in \mathbb{R}^m$ with $m < n \ \in \mathbb{R}$ and

$$h(x) = f(W_1 x + b_1),$$

where $f$ represents a non-linear activation function, e.g. sigmoid function, $W_1 \in m \times n$ a weight matrix and $b_1 \in \mathbb{R}^m$ a bias vector. The output layer decodes the latent space, i.e. hidden representation of input examples, aiming a reconstruction of $\tilde{x} \in \mathbb{R}^n$ with

$$\tilde{x} = f(W_2 h(x) + b_2),$$

where $W_2 \in n \times m$ is a weight matrix and $b_2 \in \mathbb{R}^n$ a bias vector. The training procedure minimizes the reconstruction error $\vartheta$ by finding parameters $\theta = \{W_1, W_2, b_1, b_2\}$ where

$$\vartheta(\theta) = \sum_{\mathrm{x} \in \mathrm{X}} \|x - \tilde{x}\|^2$$

## 3.3 Selecting and evaluating machine learning methods

Having introduced a variety of machine learning methods, the question that is yet to be answered remains which method to apply under which circumstances. This also includes evaluation of the model's quality or "goodness of fit". Unfortunately, there is no straightforward answer to these questions. Machine learning methods are highly problem-specific and as has been mentioned already, often deep knowledge of the data is necessary in order to decide which method to apply. Thus, there are guidelines for choosing models (such as in [113]), but these are to be understood as recommendations rather than objective rules.

Similarly, evaluating the quality of a particular model being applied to data is also a problem-specific and even an algorithm-specific matter. For supervised learning methods, model quality is ensured by separating a part of the data from the rest of the dataset and not using it when training the model. After the model has been trained on the rest of the dataset (the training set), this so-called test set is used to compare the model's prediction against the correct label [114]. Common metrics used to do this are, among others, the Mean Absolute Error (MAE), Mean Absolute Percentage Error (MAPE), and the Root Mean Square Error (RMSE) [115] (see Figure 11 for the calculation of the different error measures). This allows comparing the performance of different machine learning approaches for the same task.

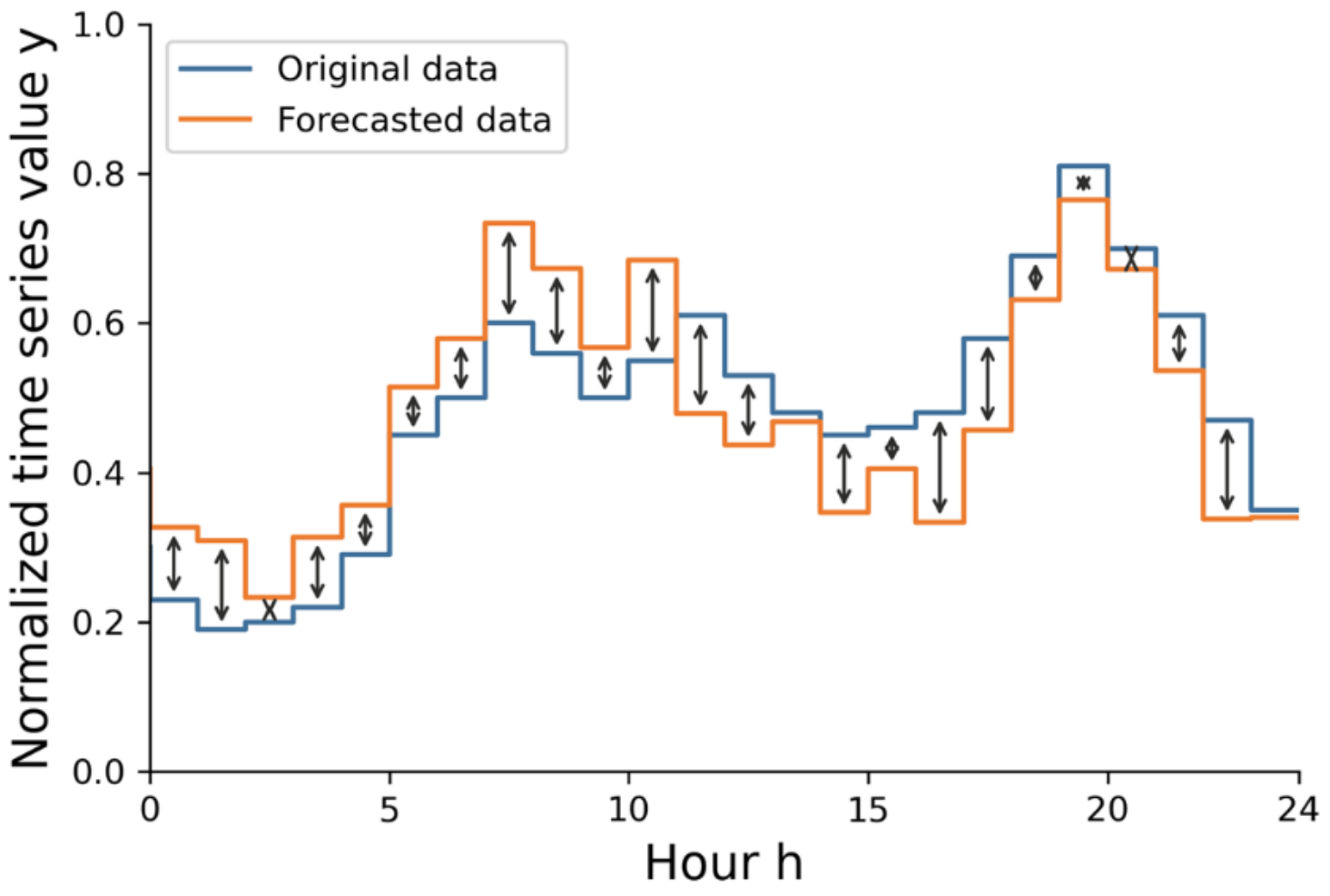

**Definition of error measures**

$$RMSE = \sqrt{\frac{1}{H}\sum_{H}(y_h - \widehat{y_h})^2}$$

$$MAE = \frac{1}{H}\sum_{H}|y_h - \widehat{y_h}|$$

$$MAPE = \frac{1}{H}\sum_{H}\left|\frac{y_h - \widehat{y_h}}{y_h}\right|$$

$y_h: acutal\ value\ for\ hour\ h$
$\widehat{y_h}: predicted\ value\ for\ hour\ h$

*Figure 11: Exemplary illustration of the deviation between actual and predicted electricity loads based on Hoffmann et al. [116].*

Ensuring the quality of an unsupervised learning approach, such as clustering or PCA, is less straightforward since these algorithms are designed for tasks where there is no correct label. For example, in clustering, a variety of generic (so-called external) evaluation metrics have been proposed, which allow comparing the results of different clustering algorithms. However, these metrics rely on external data (i.e. correct labels), which is often not available [117,118]. So-called internal metrics rely only on the information inherent in the input data and are often based on the metric the algorithm tries to optimize [117], making them algorithm-specific. This makes different clustering results difficult to compare.

Both supervised and unsupervised machine learning methods, however, share the necessity to perform so-called *hyperparameter tuning*. The various methods introduced in this paper typically require the user to set a number of parameters, the number of neurons and layers in an ANN, or the number of clusters in k-Means clustering [119]. The optimal set of hyperparameters often depend not only on the task at hand but also on the concrete dataset and the choice of hyperparameters can influence the model performance to a great extent [120]. Thus, careful choice of hyperparameters is important in order to achieve satisfactory results. Typically, hyperparameter tuning is performed manually or through heuristics, although automated approaches for this procedure are being investigated [119]

# 4 Metamodeling and design of experiments

Depending on the complexity of a simulation model and the available time span, the simulation duration can limit the scenario scope and thus the depth of analysis. A very effective method to increase the scenario scope of complex simulation models is the so-called metamodeling. To create a metamodel, information about the system behavior of the given simulation model is needed. In this section, metamodeling (see section 4.1) and a very effective method for minimizing the simulation effort to generate the required information about the system behavior, the so-called design of experiments (see section 4.2), are presented in more detail.

## 4.1 Metamodeling of simulation models

The direct use of complex simulation models for in-depth analyses is only possible to a limited extent due to long simulation durations. Metamodels are able to achieve predictions with high quality in a few milliseconds. The term metamodel and the concept goes back to the works of Blanning [121–123]. This method became more popular with the work of Kleijnen who extended it with some statistical tools [124]. These metamodels represent the system behavior of the simulation model by mapping a relationship between the input and output variables. This is realized by a mathematical approximation. Metamodels are also called approximation models, surrogate models, and response surface models [125]. Figure 12 schematically shows the procedure of metamodeling for the case of security of electricity supply assessment.

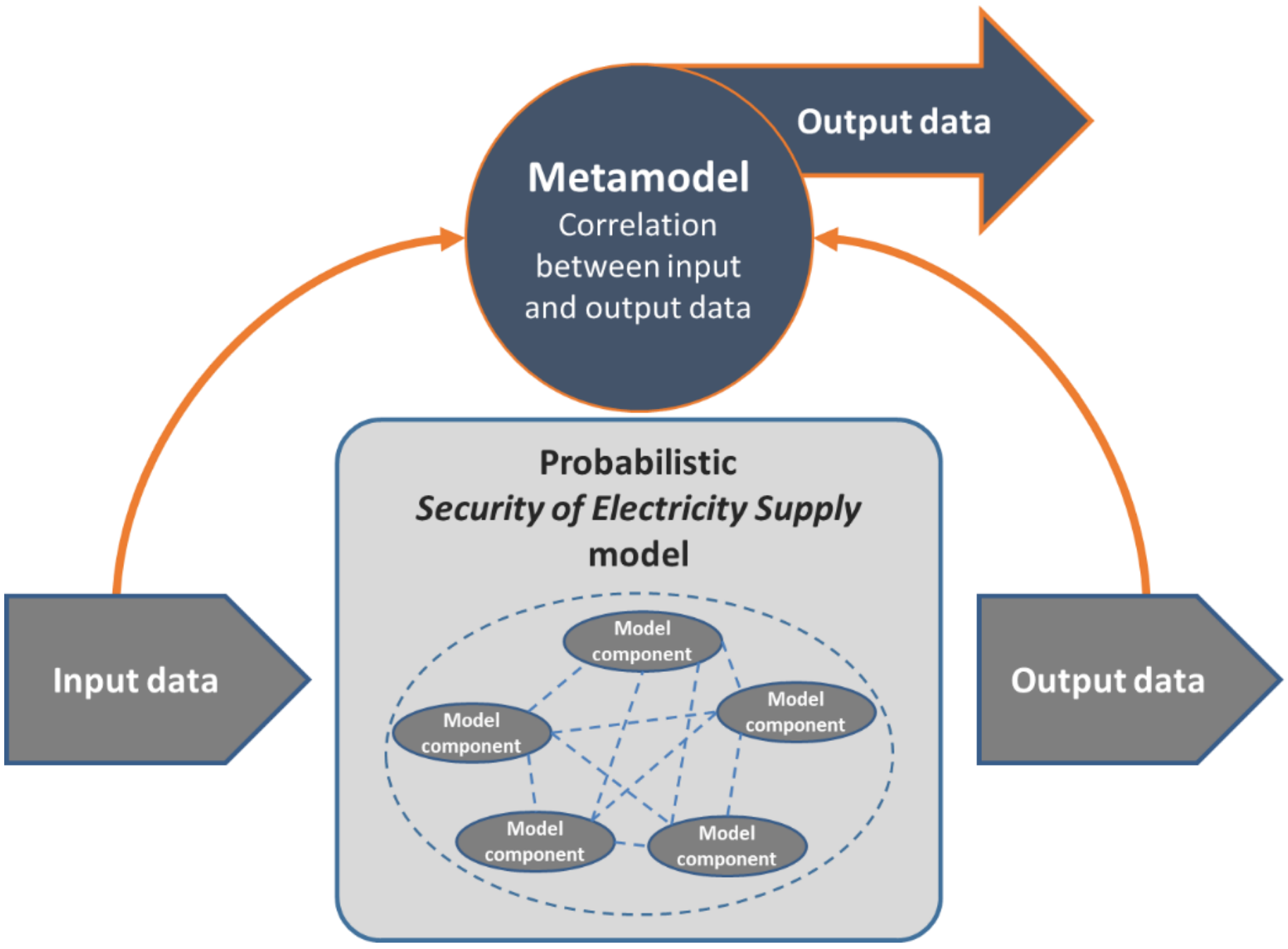

*Figure 12: Schematic representation of the basic principle of metamodeling for the case of security of electricity supply assessment.*

Metamodels are generated from real simulation data and are valid for a predefined design space. The design space represents a multidimensional structure spanned by the input data of the simulation model and that comprises the complete range of all input data combinations. The boundaries of the design space are thus defined by the minima and maxima of the input data of the simulation model. A subset of this input data is selected as the feature set and the corresponding output data (that is calculated via the simulation model) serves as the label set. Together, these selected features and calculated labels constitute the sample to which metamodel is fitted. An illustration of a design space and selected features for a metamodel is given in Figure 13.

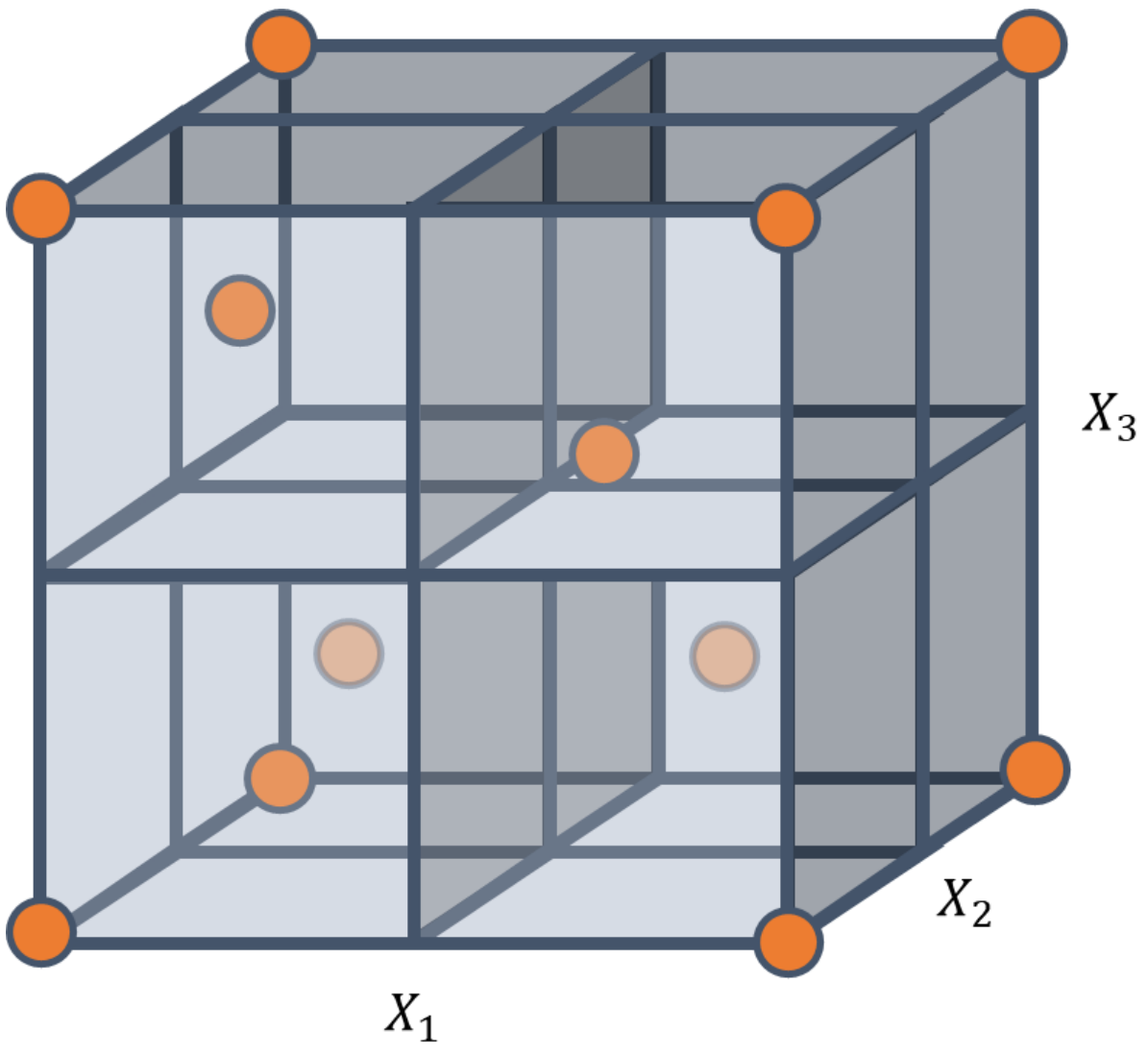

*Figure 13: Design space with selected input data combinations that serve as features for a metamodel.*

Once the metamodel has been created, the label, i.e. the output data, of any feature combination of the input data within the design space can be predicted. When selecting the feature sample from the input data, care must be taken that the amount of information gained is sufficient to represent the system behavior. At the same time, the effort to generate the information must be minimized. The most effective way to implement this is to use methods from design of experiments (s. section 4.2).

For metamodels, a variety of methods can be applied which achieve varying degrees of accuracy depending on the problem at hand. Several classical statistical methods can be used. Among them are for example linear or polynomial regression. In many use cases, simulations have a more complex nature. This often leads to nonlinear relation between the input and output data which cannot be sufficiently approximated by classical statistical methods. For these cases, machine learning methods can be applied (see section 3.1 for details on a selection of methods).

The choice of the approximation method and the experimental design depends on the problem at hand and the optimal choice, in most cases, cannot be determined a-priori (see section 6). For this reason, testing and validating different approaches is a very important part of metamodeling. To this end, many evaluation variables, such as the coefficient of determination $R^2$, are available from statistics. Decisive for the testing of different metamodels is the quality of the validation data. These samples must be independent of the training data, which were used for the metamodeling itself. Furthermore, test data must be evenly distributed over the design space, so that the evaluation of the predictive quality is representative of the entire design space. [126]

## 4.2 Design of experiments for effective scanning of the design space

Design of experiments (DOE) is a method for efficient planning and designing experiments. Depending on the complexity, experiments can be costly and lengthy. As a consequence of a limiting budget or given time span it is rarely feasible and probably never reasonable to carry out a series of experiments in an unplanned manner. This is particularly the case for models that are intended to depict complex relationships of the energy system. To get the most information out of the system under different limiting circumstances DOE is applied.

Statistical design of experiments was developed already in the early 20th century and was originally developed for real experiments [127]. The methodology was then later adapted for the statistical DOE [128,129]. In the following, we will give a short overview of the concept of DOE for the case of metamodeling simulation models. For an extensive overview of the DOE methodology and a variety of experimental designs, see Montgomery [130] and Siebertz et al. [126].

### 4.2.1 Methods for setting up experimental designs

The most important part of the DOE is the selection of the appropriate design. In this subsection, a short overview of a selection of designs will be given. For reasons of brevity, we focus on describing basic features of the respective methodologies. In addition to these designs, there are several other important designs. Among them are screening designs, Box-Behnken designs, and the quasi-Monte Carlo method.

**Full-factorial and fractional factorial designs:** If a linear relationship is suspected, the most important experimental designs are full-factorial and fractional factorial experimental designs according to the Yates standard [131]. The term factor is used synonymously with feature in the context of DOE because of its original application to real experiments. Full-factorial and fractional factorial experimental designs contain the extreme value combinations on the vertices of the factor space (see Figure 14). The number of factor combinations results from the number of variable factors $k$. A full-factorial experimental design results in $2^k$ combinations and $2^{k-p}$ for fractional factorial experimental designs where p is the number of commingling. In the case of fractional factorial experimental designs, information is lost due to the commingling of (supposedly) negligible interactions. These experimental designs are only used if it is evident that the commingling variables are negligible or if the number of experiments should be reduced for efficiency purposes.

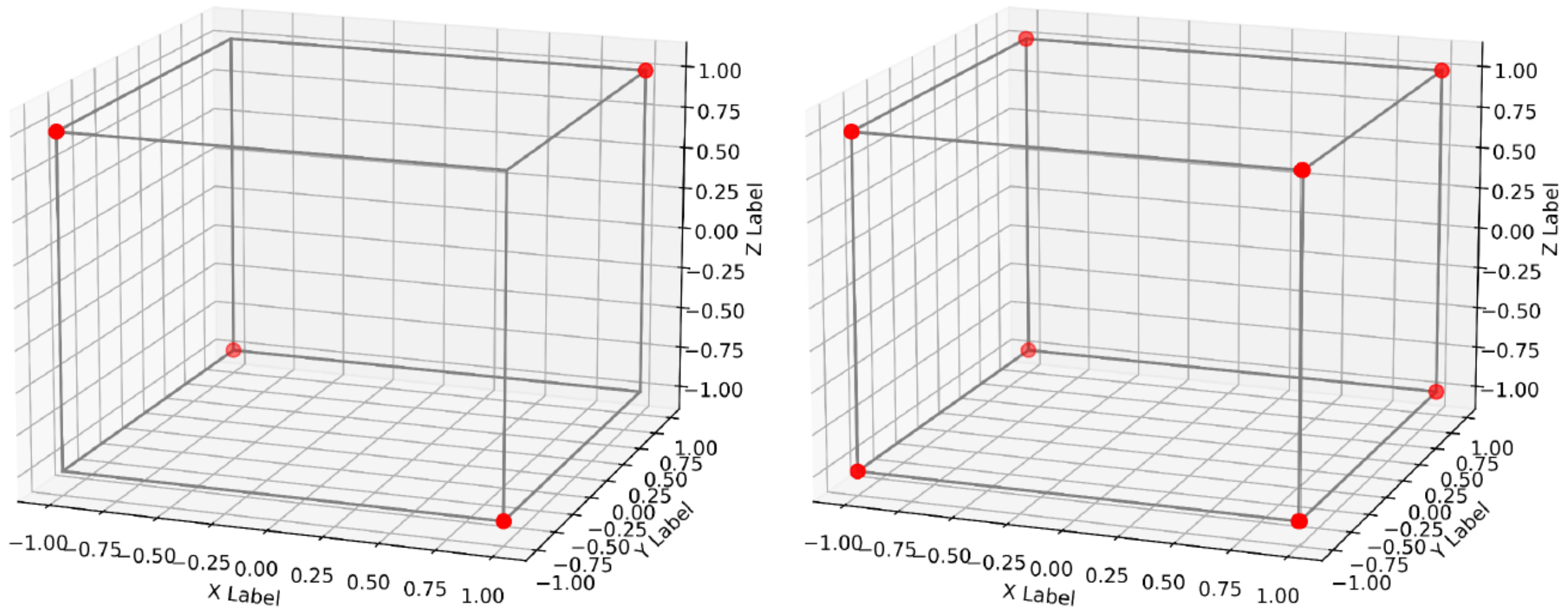

*Figure 14: Full-factorial (left) and fractional factorial (right) experimental design (k=3)*

**Central composite designs:** If a non-linear relationship is suspected, the above experimental designs are no longer sufficient. It becomes necessary to extend the ability of the metamodel to also consider the quadratic terms of the main effects. An experimental design often successfully used is the so-called central composite experimental design (CCD). The CCD can be thought of as an extension to the formerly described designs: A (fractional) factorial experimental design is extended by so-called "star" points (see Figure 15) as well as center points. The additional points added to the experimental design allow for evaluating potential quadratic effects. The star points extent the (fractional) factorial design space in most cases and are then called circumscribed. Choosing the distance of the star points to the (fractional) factorial design space is given by choosing a desired statistical property the final design should have.

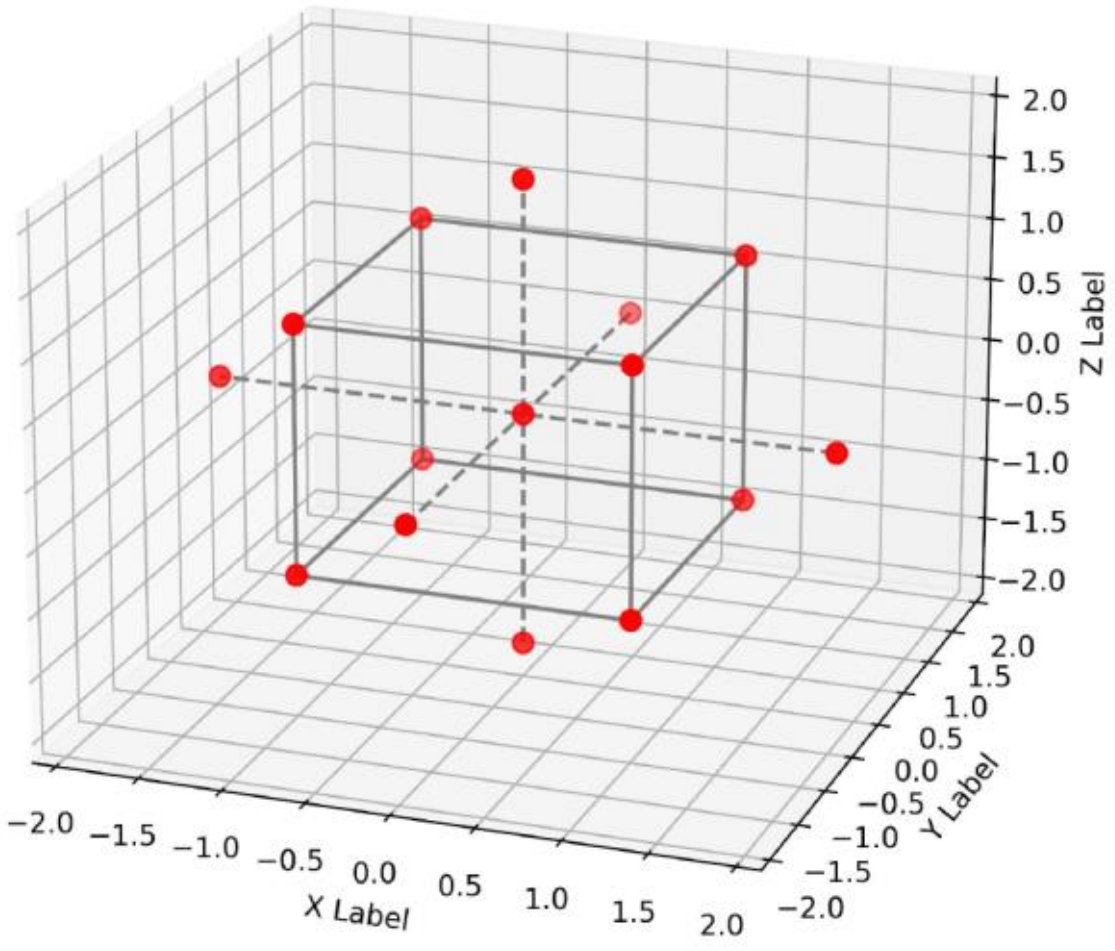

*Figure 15: Central composite experimental design (CCD) (k=3)*

There is also a subtype of this design, the so-called face-centered-central-composite design. In this design, the factor combinations of the "star-shaped" design lie on the plane spanned by the corner points of the full/partial factorial experimental design. However, this worsens the representation of the non-linear behavior. Therefore, it is only used if a factor can only be varied as an integer.

**Latin-hypercube design**: Another class of designs that are increasingly used in non-linear contexts, especially when dealing with computer-aided experiments (CAE), are the space-filling designs. One example of those designs is the so-called Latin-hypercube design (LHD) [132,133].

For the LHD, the factor combinations are determined using some sort of stochastic procedure, for example by using a randomly permutated array to construct the full design by a given logic [134]. This method is resembling the Monte Carlo method, where the factor combinations are also determined randomly. In contrast to this, the Latin Hypercube uses a methodology to ensure uniform coverage of the entire multidimensional design space (see for an exemplary LHD). If an LHD is well constructed, the variance of the global mean will be significantly lower than when using a random Monte Carlo field with the same number of test points.

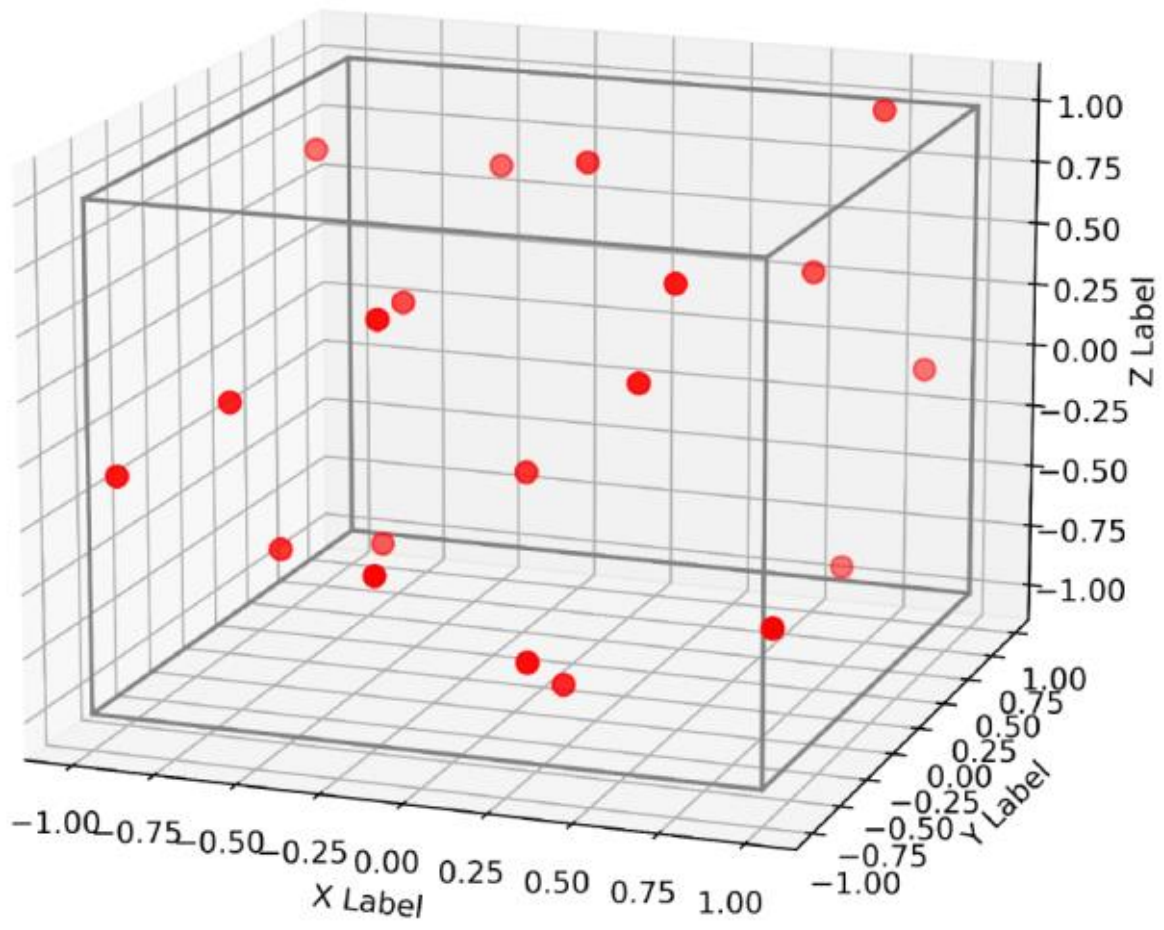

*Figure 16: Latin-hypercube design (k=3)*

Basically, in an LHD the design space is first divided into zones. From this zoned design space, a random factor combination is then determined in each zone. A uniform and correlation-free coverage of the factor space is not automatically ensured. For this, further methods such as orthogonal designs or rather space-filling design would have to be applied [134]. There are a variety of possible methods for constructing an LHD. The selection is strongly dependent on the problem at hand. Recommended approaches are described for example by Moon [135] and Dash et al. [136]. A good overview of space-filling methods and publications is provided by van Dam et al. [137].

### 4.2.2 Current developments in design of experiments

DOE has been used successfully for a long time and the individual methods have a high standard. A very large potential for improvement and further development lies in a subarea of machine learning for whose application design of experiments is essential, the so-called area of active learning also known as query learning [138,139]. In the statistical literature, this application area is also called optimal or adaptive DOE.

Settles describes the basic idea of active learning as follows: “[…] that a machine learning algorithm can achieve greater accuracy with fewer training labels if it is allowed to choose the data from which it learns”. The goal is to minimize the number of factor combinations while achieving a certain forecast quality in combination with metamodeling. Settle defines the term active learning in his work as follows: “Active learning systems attempt to overcome the labeling bottleneck by asking queries in the form of unlabeled instances to be labeled by an oracle (e.g., a human annotator). In this way, the active learner aims to achieve high accuracy using as few labeled instances as possible, thereby minimizing the cost of obtaining labeled data”. For the case of the assessment of security of electricity supply, this oracle is the probabilistic simulation model. Settles divides active learning into three main scenarios membership query synthesis [140], stream-based selection sampling [141,142], and pool-based sampling [143]. Figure 17 shows an illustration of these concepts.

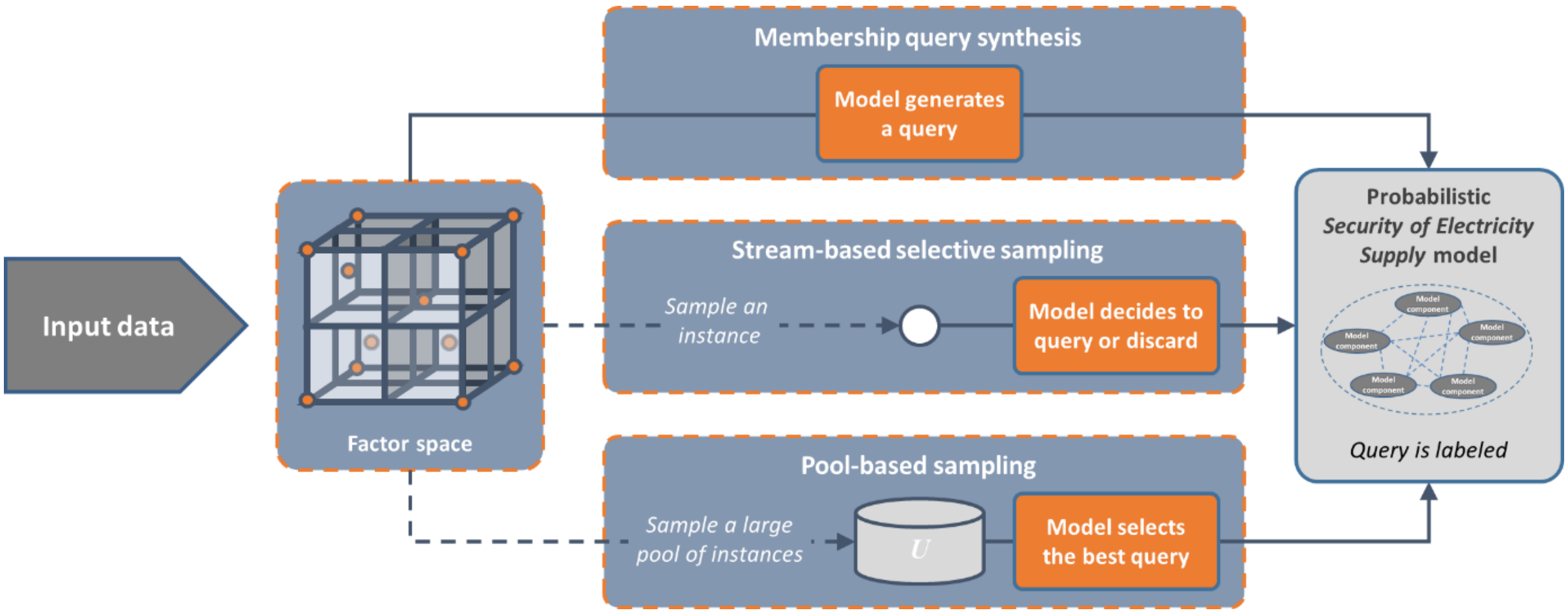

*Figure 17: Schematic representation of the three main scenarios of active learning in the context of the assessment of security of supply. Own representation based on Settles [138].*

According to Settles [138], there are a variety of different approaches to active learning. For example, one approach to metamodeling a simulation model using active learning is to add iterative factor combinations to a baseline experimental design. In each iteration, metamodeling and validation are repeated until the desired prediction quality is achieved. The difficulty is to scan the experimental space as effectively as possible and to select the factor combination that provides the greatest possible additional information benefit about the system behavior. For the selection of these factor combinations, there are different basic strategies on which also more current approaches are based. Among others are *uncertainty sampling* [143], *query by committee* also known as *ensemble-based strategy* [144], and *expected error reduction* [145].

Current work is focused on the further development of approaches for different application areas and optimization for different metamodeling methods. For example, neural networks with active learning still have a lot of potential. In particular, with the increased use of deep learning methods in recent years, a large research field of deep active learning has emerged. The combination of active learning and deep learning poses some challenges. In contrast to statistical approximation methods, the strengths of deep learning methods do not lie in showing where the uncertainty in the prediction is large. Also, iterative approaches are very computationally intensive since the network must be retrained in each iteration. A very first general overview of this very broad field is given by Ren et al. [146] and Liu et al. [147].

## 4.3 Combine metamodeling with design of experiments

A possible methodical approach to metamodeling simulation models is demonstrated in the work of Reich et al. [148]. This approach is done using a model for the simulation of an energy supply systems but can also be applied to other simulation models. The authors conclude that using an LHD to sample the information to train an artificial neural network is the best approach for approximating the response of the analyzed energy supply system. Furthermore, LHDs and ANNs can be used more flexibly and can thus be better adapted to the problem at hand.

For the presented approach, six steps are defined by the authors: Problem definition, defining the design space, developing experimental designs, developing approximation models, comparison, and validation as well as the system analysis (see Figure 18).

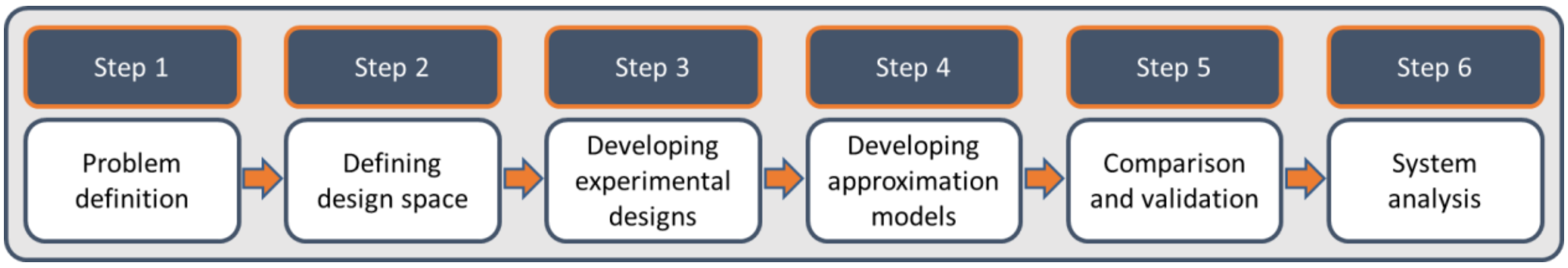

*Figure 18: Methodical approach to the metamodeling of a computer simulation based on Reich et al. [148].*

In the following, the individual steps of the approach are presented only briefly and in generalized form.

**Problem definition:** Depending on the problem, factors are determined which could have a decisive influence on the system behavior of the simulation model.

**Defining design space:** In the second step, the design space is defined for which the later metamodel is valid. For this purpose, the boundaries of the factors selected in step one are set depending on the problem.

**Developing experimental designs:** To reduce the required factor combinations, the third step is the selection of the experimental design and transfer to the created design space.

**Developing approximation models:** Once the simulation model has been used to generate the system information for the factor combinations of the experimental design, approximation methods can be used to create a metamodel.

**Comparison and validation:** To evaluate how well the metamodel can approximate the systems response, the prediction quality must be evaluated. The meaningfulness of the prediction quality depends decisively on the quality of the test points. High-quality test points have two characteristics. Firstly, the test points have as uniform a distribution as possible in the design space and, secondly, they were not used to create the metamodel.

**System analysis:** With a metamodel capable of representing system behavior, a variety of possible analysis options are possible. These are among others, the evaluation of the effects of individual input variables on the output variables, sensitivity analysis, large-scale scenario analysis, and multi-criteria optimization.

# 5 Field of application: AI-based forecasting methods for the assessment of the security of electricity supply

In this section, we will introduce a selection of applications for machine learning in energy system analysis with a focus on the assessment of the security of electricity supply. We do not claim the list to be exhaustive but rather selected those applications that we regard as most relevant. For each field of application, we conducted a systematic literature research.

A general distinction of the structure of forecasting models can be seen in Table 3. The information provided is a simplified representation of model structures and does not account for feedback or other more complex variants.

*Table 3: Categorization of the general structure of forecasting models*

| | | **Temporal overlap of input data and output data** | | |
|---|---|---|---|---|
| | | No overlap, without foresight | No overlap, with foresight | Overlap, with foresight |
| **Input feature space** | Univariate | 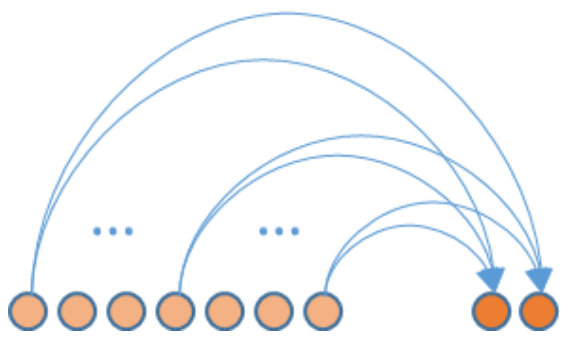 | -/- | -/- |
| | Multivariate | 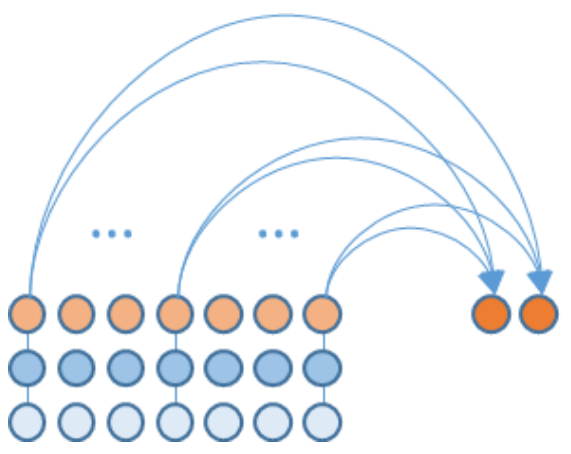 | 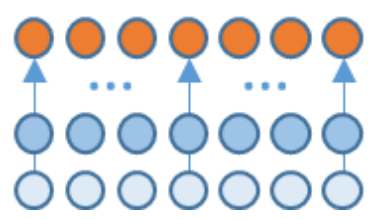 | 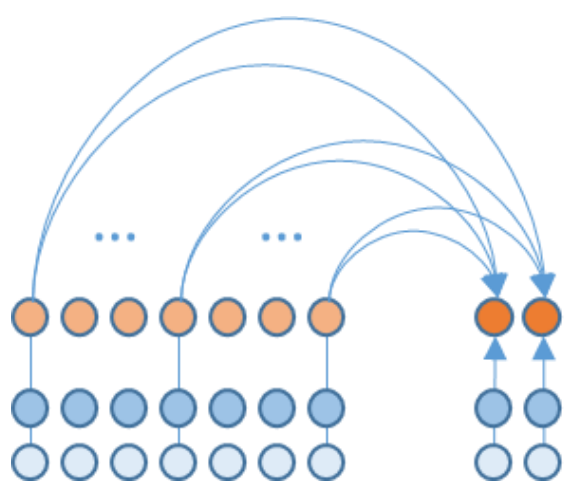 |

A univariate forecasting model is based entirely on historic information on the output feature. In contrast, multivariate models use additional information as input features that correlate with the output feature. Multivariate models can be further distinguished between using information predictions for the future values of input features or not. In addition, a hybrid model that uses both, historic information and future predictions of input features can be constructed. In this case, further prediction models are used for forecasting input feature values.

## 5.1 Forecasting electricity load profiles

Forecasting electricity load profiles can be categorized into four different time horizons shown in Table 4:

*Table 4: Definition of four different time horizons for load profile forecasting [7].*

| Abbreviation | Forecasting duration | Time interval |
|---|---|---|
| VSTLF | Very short-term load forecast | Less than 30 min. |
| STLF | Short-term load forecast | 30 min. to 1 day |
| MTLF | Medium-term load forecast | 1 day to 1 year |
| LTLF | Long-term load forecast | >1 year |

Electricity load forecasts are used in different fields of application [149]. For assessing the security of electricity supply, long-term forecasts are used to plan and build infrastructure for a secure provision of electricity. Electricity load profiles are affected by a multitude of factors, such as weather, economic factors, or the portfolio of end-user technologies [12].

**Forecasting data:** Electricity load profiles are given as time series with varying temporal resolutions. Electricity loads can be provided as an aggregated information, usually from minute to hourly averages [150]. Average values have the advantage that, aggregated, they represent total annual consumption. However, due to the aggregation, peak loads are usually neglected that would be better represented by maximum instead of average values. In turn, using the maximum function for aggregating temporal data would overestimate the total annual consumption. Hybrid solutions using averaging as an aggregation method and adding peak load values are used to compensate for the individual disadvantages of the aggregation methods. Electricity loads can be subject to different aggregation levels [151]:

- Sectoral aggregation: Total load profiles vs. sectoral load profiles
- Regional aggregation: No regional differentiation vs. including regional identifiers
- Temporal aggregation: Time series data in sub-hourly, hourly, daily, or weekly resolution

In principle, the less aggregated the data, the more accurate the representation of the real system. However, this principle does not apply if the quality of the data sample is low. The level of aggregation of the data should, therefore, be chosen so that the accuracy of the data at the chosen resolution meets the requirements. This means that data may need to be aggregated.

**Feature selection:** The main features influencing electricity loads are calendrical information, meteorological (temperature and weather) data, and economic factors (such as future prices or prices of other energy carriers) [152]. Features, as well as the forecasting data, can vary in terms of sectoral, regional, and temporal aggregation. Table 5 shows a selection of studies on medium- and long-term forecasting and the features used in the forecasting models.

*Table 5: Comparison of features used for predicting electricity loads based on selected studies*

| Feature \ Study | Melodi et al. [153] | Sangrody et al. [154] | Yasin et al. [155] | Matsuo and Oyama [156] | Behm et al. [7] | Dai and Zhao [157] |
|---|---|---|---|---|---|---|
| **Historic load** | X | X | X | X | | |
| **Calendrical information** | | X | | X | X | X |
| **Air temperature** | | | X | X | X | |
| **Wind speed** | | | X | | X | |
| **Irradiation** | | | | | X | |
| **Relative humidity** | | | X | | | |
| **Weather classification** | | | | X | | |
| **Heating day information** | | X | | | | |
| **Cooling day information** | | X | | | | |
| **Price data** | | | | | | X |
| **Gross domestic product** | X | | | | | |
| **Population** | X | | | | | |

**Prediction models:** Figure 19 shows the results of the systematic literature review on the development of electricity load forecasting models within the Scopus database.[7]

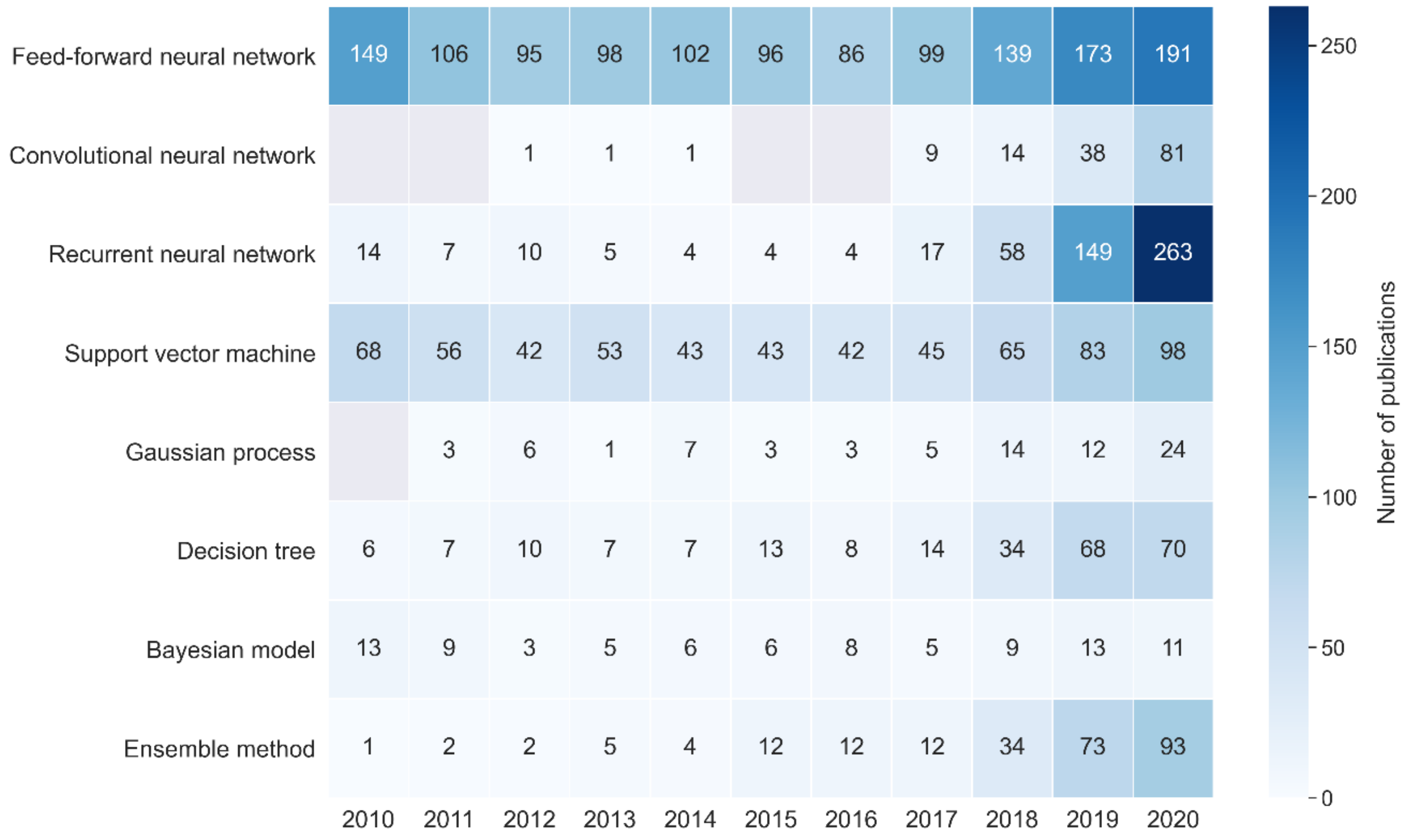

*Figure 19: Number of scientific articles found on load forecasting in energy-related journals in the Scopus database from 2010 to 2020, categorized by AI-based methodology.*

Neural Networks are the most widely used AI methodology for predicting electricity load time series. Studies using FFNN are published at numbers between ~80 and ~200 per year. Studies using CNNs and RNNs have become increasingly popular since 2018, with RNNs being used even more frequently than FFNNs in 2020. In the field of RNN, LSTM neural networks dominate and are used in the majority of studies. Support vector machines are constantly used in ~40 to ~100 studies per year, making them a popular alternative to neural network approaches. Finally, the number of studies based on decision trees and ensemble methods has recently increased significantly since 2018.

## 5.2 Forecasting renewable feed-in profiles

In addition to the demand side, there is also uncertainty on the supply side, driven in particular by the expansion of renewable energies. This is the most critical scheduling input, as both situations of oversupply and undersupply are possible. In the context of assessing the security of electricity supply, undersupply is of particular relevance [12]. However, the inclusion of situations of oversupply in the analysis becomes more important the more storage facilities in the system can absorb this energy and make it available again at times of undersupply [158]. Similar to forecasting electricity load profiles, the time horizon of the forecast can be used as an initial distinction of the fields of application:

*Table 6: Generation forecast methods and applications [159,160]*

| Time horizon | Methods | Key applications |
|---|---|---|
| 5 – 60 min. ahead | Very short-term forecast | Regulation, real-time dispatch, trading, market- |

[7] For details on the search term specification, see Table 12 in the appendix.

| | | |
|---|---|---|
| | | clearing |
| 1 – 6 hours ahead | Short-term forecast | Scheduling, load following, congestion management |
| Day(s) ahead | Medium-term forecast | Scheduling, reserve requirement, trading, congestion management |
| Weeks or more ahead | Long-term forecast | Resource investment planning (generation, network), contingency analysis, maintenance planning, operation management |

For the field of application portrait in this paper, long-term forecasting is again the relevant time horizon. Renewable feed-in profiles combine the availability of the power plant with a capacity credit due that is based on weather forecasts [158]. We will address the forecasting of non-availabilities in section 5.3 in more detail. Forecasts of renewable feed-in profiles can now focus on the solely weather-dependent aspects or a combination of the two.

**Forecasting data:** Renewable feed-in profiles are given as time series with varying temporal resolutions. Similar to electricity load profiles, they are given as an aggregated information providing, e.g., minute or hourly averages [161]. In addition to average values and in contrast to electricity load profiles, it is not maximum but minimum values that are of interest for assessing the security of electricity supply in order to perform a robust analysis that can also map extreme events. Renewable feed-in profiles can be subject to different aggregation levels:

- Technological aggregation: E.g. aggregating or disaggregating rooftop PV and ground-mounted PV
- Regional aggregation: No regional differentiation, including regional identifiers, or per unit
- Temporal aggregation: Time series data in sub-hourly, hourly, daily, or weekly resolution

**Feature selection:** The main features influencing (aggregated) renewable feed-in profiles are historical meteorological data, numerical weather predictions (NWP), and information about the surroundings to estimate the effects of shadows and wind shadows [162]. Table 7 shows a selection of studies on wind power feed-in forecasting and the features used in the forecasting models.

*Table 7: Comparison of features used for predicting wind power feed-in based on selected studies*

| Study / Feature | Chen and Folly [163] | Xiaoyun et al. [164] | Pelletier et al. [165] | Bilal et al. [166] | Shahid et al. [167] | Nazaré et al. [168] |
|---|---|---|---|---|---|---|
| **Wind speed (at different hub heights)** | X | X | X | X | X | X |
| **Wind direction** | X | X | X | X | X | |
| **Air temperature** | X | X | | X | | X |
| **Zonal and meridional flows** | | | | | X | |
| **Air density** | | X | X | | | |
| **Air pressure** | X | X | | | | X |
| **Relative humidity** | X | | | X | | X |
| **Turbulence intensity** | | | X | | | |
| **Wind shear** | | | X | | | |
| **Yaw error** | | | X | | | |
| **Solar irradiation** | | | | X | | |

**Prediction models:** Figure 20 shows the results of the systematic literature review on renewable feed-in forecasting.[8]

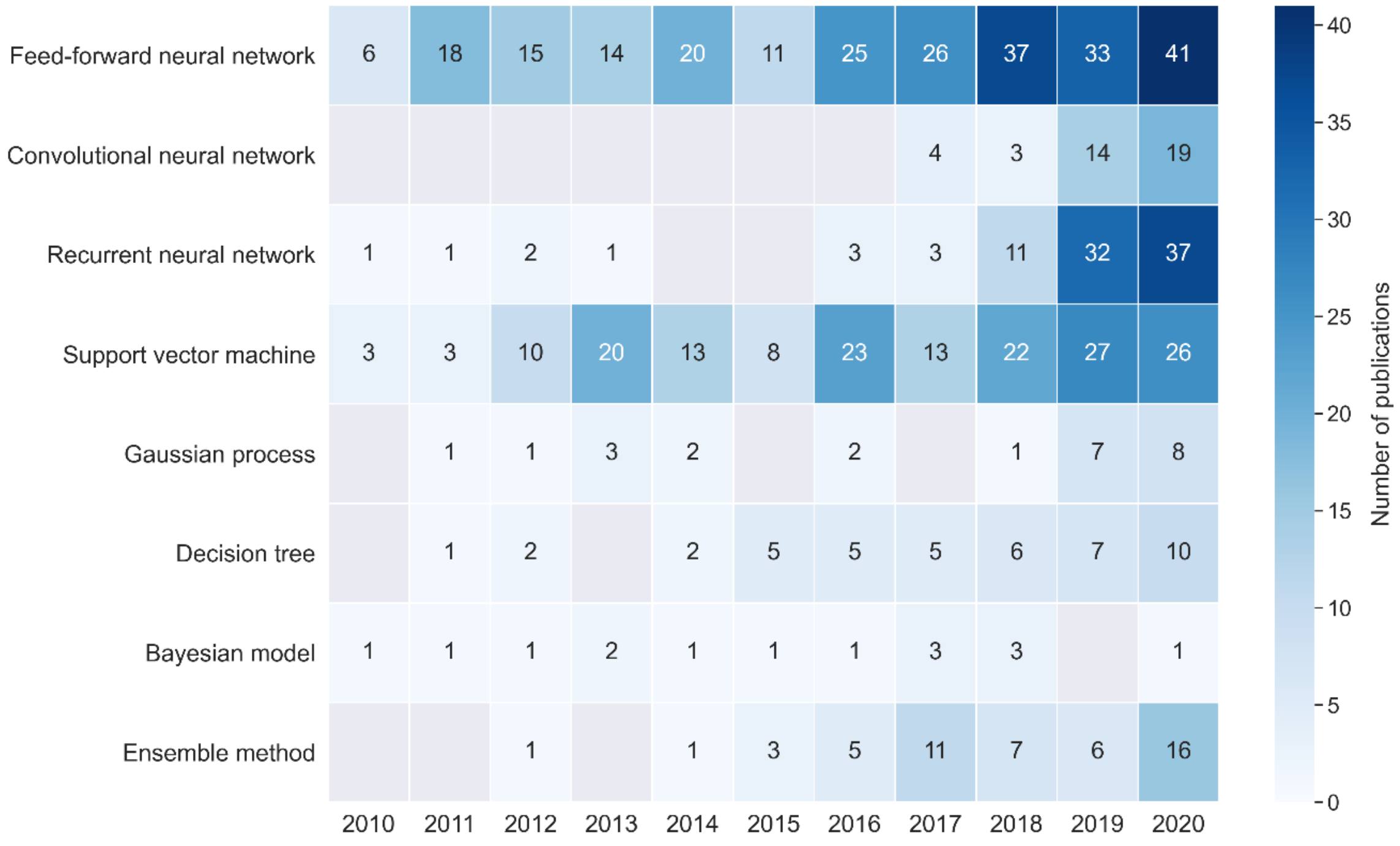

*Figure 20: Scientific journal articles on renewable feed-in forecasting in energy-related journals in the Scopus database from 2010 to 2020, categorized by AI-based methodology.*

Similar to studies on predicting electricity load profiles, FFNNs are again the most widely used AI methodology for predicting renewable feed-in time series. CNNs and RNNs have drastically increased in popularity since 2018. While support vector machines have a constant number of publications of ~20, other methods such as decision trees, Bayesian models, or ensemble methods are not yet used as frequently for predicting renewable feed-in time series.

## 5.3 Forecasting (non-)availabilities

A major uncertainty in assessing security of electricity supply is the availability of generation capacities and network components. Components in the energy system can be non-available due to planned and unplanned outages [12]. Planned non-availabilities are known in advance and are usually due to scheduled (e.g. annual) maintenance. Unplanned non-availabilities are not known in advance, are subject to a much more random distribution than planned non-availabilities, and can be due to malfunctioning or uncontrollable external factors such as extreme weather conditions. Both types of events have a major impact on supply security [169]. Due to the differences in their distribution and influencing factors, they are usually modeled and predicted separately [170].

The application field of the availability of power plants for machine learning-based methods is therefore twofold. For planned unavailability, scheduled maintenance cycles need to be predicted [12]. Insights from predictive maintenance can be transferred to improve predictions on maintenance schedules [171]. In contrast, in the case of unplanned unavailability, factors such as complex thermodynamic and the security of supply of fuels need to be modeled [172]. This imposes fundamentally different requirements on predictive models.

[8] For details on the search term specification, see Table 13 in the appendix.

Another challenge that stems from the systemic perspective in the assessment of security of electricity supply is, that no further operational data from specific sites is available. This is due to the fact that the entirety of power plants must be depicted, with various operators and at various locations. Forecasting models, therefore, have to rely solely on external factors such as weather conditions that can be monitored and predicted.

Finally, a distinction needs to be made between modeling non-availability of components independently or (1) considering common-mode situations and (2) considering time-dependence [173–175].

**Forecasting data:** Component unavailability can be available as binary information, multiple discrete states, or continuous availability levels [170,173]. Binary data indicates whether the component is available or not while discrete and continuous data additionally show the share of non-available capacity. Data on non-availability are subject to different aggregation levels:

- Technological aggregation: Total capacity, capacity per generation technology, or capacity per generation unit
- Regional aggregation: No regional differentiation or including regional identifiers
- Temporal aggregation: Annual data (availability factor), data for certain points in time (e.g. during annual peak load), or time series data (e.g. hourly resolved)

**Feature selection:** We assume the main features influencing the availability of generation capacities to be calendrical information, technology-specific data, weather and further environmental data, price data, and load data. Table 8 shows a selection of studies on (non-)availability forecasting for thermal power plants and the features used in the forecasting models. The studies listed do not exclusively apply machine learning-based methods as the number of such studies found in the literature research is too low at the time the search was conducted. We assume that independent of the methodological approach, the listed features can serve as a good starting point for constructing machine learning-based models. Interestingly, price data (forward or spot prices) are not found as explanatory variables in literature.

*Table 8: Comparison of features used for predicting (non-)availabilities of thermal power plants based on selected studies. Additional information on the methodological approach and the dependencies mapped in the model is provided.*

| Study / Feature | Koch and Vögele [176] | Gils et al. [170] | Yuyama et al. [177] | Murphy et al. [174] | Nolting et al. [12] | Hundi and Shahsavari [172] | Malladi et al. [173] |
|---|---|---|---|---|---|---|---|
| **Method applied** | Analytical model | Mean-reversion Jump-diffusion model | Lognormal and Weibull hazard models | Non-homogeneous Markov and Logistic regression model | Sliding window technique | Linear regression, SVM, random forests, ANN | Continuous time Markov chain with dependence sets |
| **Non-availability category** | Un-planned | Planned, unplanned | Unplanned | Unplanned | Planned | Unplanned | Unplanned |
| **Number of plants modeled** | Single | Multiple | Multiple | Multiple | Multiple | Single | Multiple |
| **Time dependence** | | | | X | X | | X |
| **Common-mode failures** | | | | X | | | X |
| **Asset type** | X | X | X | X | X | X | X |
| **Asset age** | | | X | X | | | |
| **Asset size** | X | | X | X | X | | |
| **Calendrical information** | | X | X | | | | |
| **Air temperature** | X | | | X | | X | |
| **Air pressure** | | | | | | X | |
| **Relative humidity** | X | | | | | X | |
| **Stream temperatures or levels** | X | | | | | | |
| **Environmental regulations** | X | | | | | | |
| **System load** | | | | X | | | |
| **Other** | X | | | | | X | X |

**Prediction models:** Figure 21 shows the results of the systematic literature review on the non-availability of generation capacities.[9]

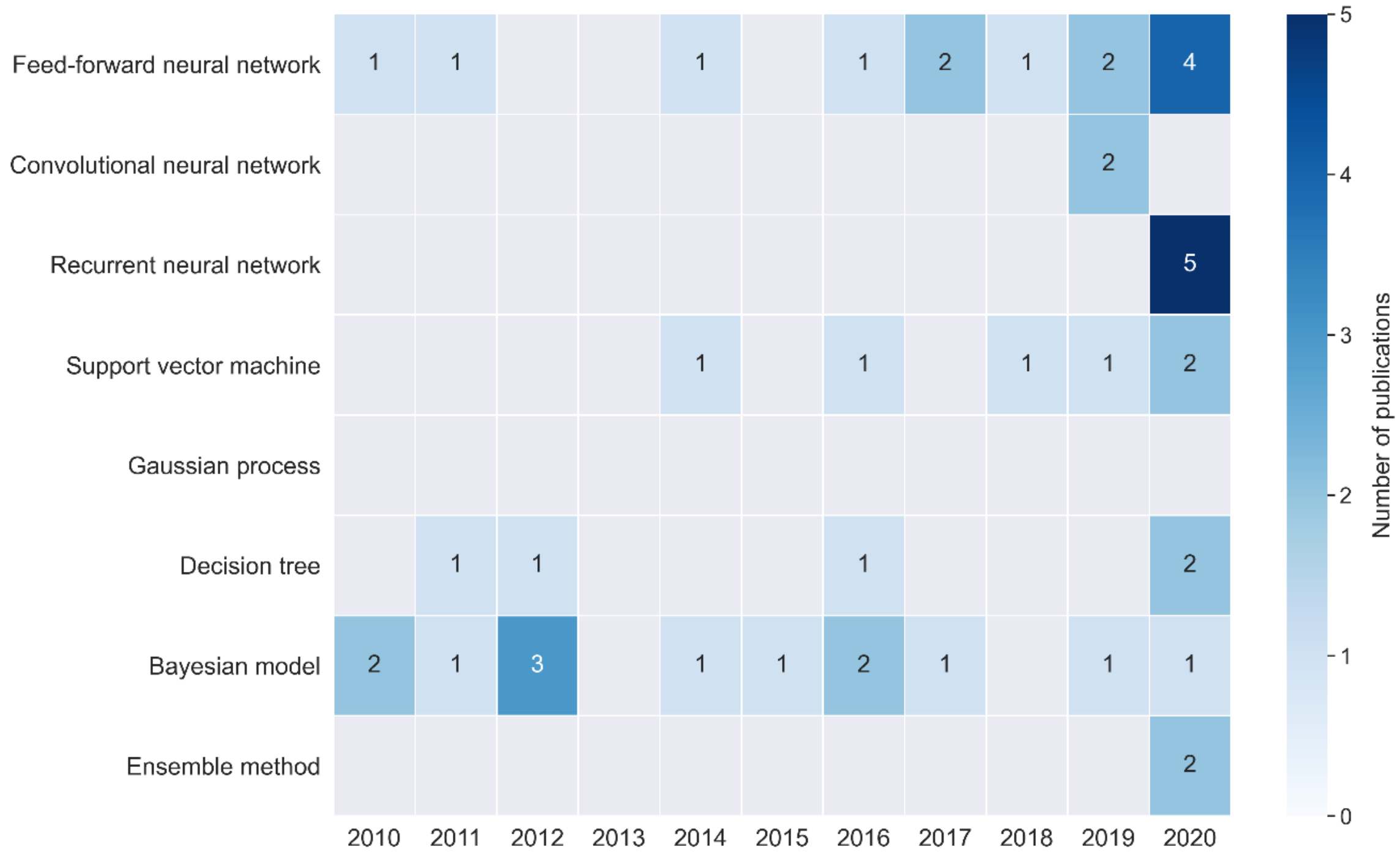

*Figure 21: Scientific journal articles on predicting (non)-availability of power plants in energy-related journals in the Scopus database from 2010 to 2020, categorized by AI-based methodology.*

The field of prediction models for the (non-)availability of power plants using machine learning is not yet established on a large scale. Neural networks are the most relevant method applied and numbers are rising over the last years. Support vector machines and Bayesian models are used in few works. Looking into the few available studies, the most cited works rely on sensor data [178–180]. As sensor data is usually used for forecasting individual plant outages, we conclude that on a system level, the application of machine learning-based methods for forecasting (non-)availability of power plants is a major research gap.

## 5.4 Forecasting storage operation

In traditional energy systems, storage systems were mainly large-scale central units such as pumped hydro storage power plants. Decentralization of the energy supply increases the number and variety of storage facilities considerably [181]. Also, new central storage technologies may be deployed in the future such as compressed air electric storages (CAES) [182]. Small and large storage units will operate differently in energy markets as large units may have an impact on the market price (price makers) while small ones do not (price takers) [183]. The increasing number and capacity of energy storage systems together with their different operating strategies, the decline in controllable power plants, and the expansion of volatile renewables increase the importance of storage systems for security of electricity supply [184]. As a result, forecasting the operation of storage is becoming increasingly important.

[9] For details on the search term specification, see Table 14 in the appendix.

Unlike thermal and renewable power plants, storage systems do not necessarily operate in line with overall systemic goals such as security of supply. Storage operators may withhold stored energy or create additional electricity demand for self-interested reasons. Rationales for this include arbitrage opportunities, extending the life of their assets, or maximizing self-consumption [185]. Large energy system models that incorporate storage alongside power plants and minimize total system costs have difficulty representing these behaviors endogenously [186]. To overcome this problem, model coupling approaches can be applied. Submodels that take the perspective of the storage operators can then represent the storage operation, which is, for example, iteratively fed into a larger system model. Such submodels could again apply cost-minimization methods or depict storage operation by applying machine learning-based methods that learn such behavior from historic data.

**Forecasting data:** Storage operation will ultimately be needed in the same temporal resolution as other temporal input data that is used for the assessment of the security of electricity supply. Data on storage operation is subject to different aggregation levels:

- Technological aggregation: Total storage dispatch and state of charge (SoC), per storage technology, or per storage unit
- Storage-specific information: Storage dispatch and/or SoC
- Regional aggregation: No regional differentiation or including regional identifiers
- Temporal aggregation: Time series data in sub-hourly, hourly, daily, or weekly resolution

Information on storage operation can be based on historic data or simulations. Scapino et al. [187] use a physics-based model to simulate a storage system and generate the forecasting data for the prediction model (this approach belongs to the field of metamodeling, see sections 4 and 6).

**Feature selection:** The main features influencing storage operation are besides technical characteristics price and other market information, weather data, and load and generation data. The studies listed do not exclusively apply machine learning-based methods as the number of such studies found in the literature research is too low at the time the search was conducted. We assume that independent of the methodological approach, the listed features can serve as a good starting point for constructing machine learning-based models.

*Table 9: Comparison of features used in studies for predicting storage operation*

| Feature \ Study | Nojavan et al. [188] | Zhour et al. [189] | Henri and Lu [190] | Mousavi et al. [191] |
|---|---|---|---|---|
| **Method applied** | Optimization | Optimization | Neural networks, random forest | Optimization and neural network (metamodeling) |
| **Storage technology** | Compressed air storage | Generic storage system | Combined PV and battery storage system | Pumped hydro storage (in an island system) |
| **Technical characteristics** | X | X | X | X |
| **Calendrical information** | | | X | |
| **Precipitation** | | | | X |
| **Air temperature** | | | X | X |
| **Relative humidity** | | | | X |
| **Wind speed** | | | | X |
| **Radiation** | | | | X |
| **Electricity price** | X | X | X | |
| **Gas price** | X | | | |
| **Electricity load** | | X | X | |
| **Renewable feed-in** | | X | X | |
| **State of charge** | X | X | X | X |

**Prediction models:** Figure 22 shows the results of the systematic literature review on predicting storage operation.[10]

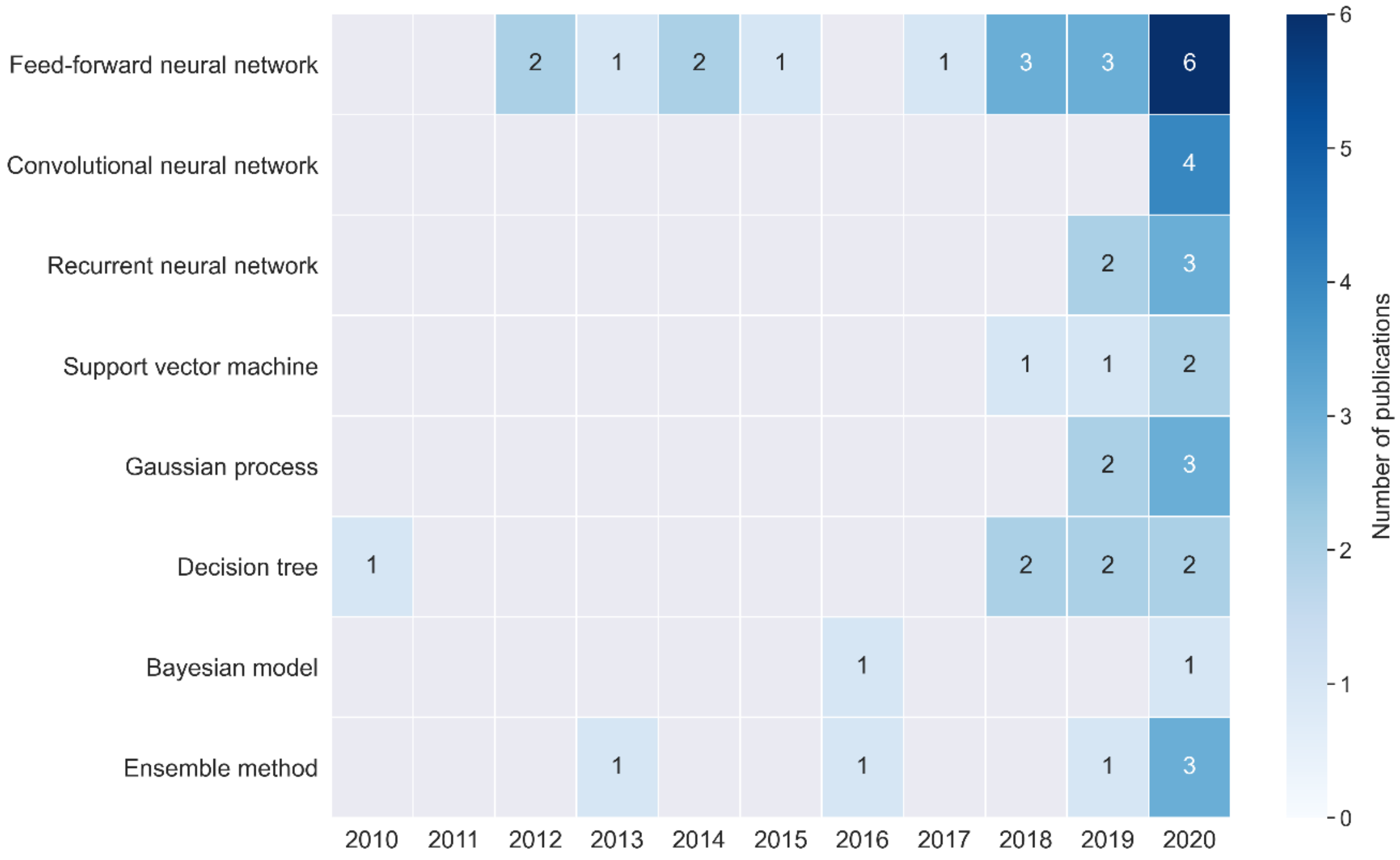

*Figure 22: Scientific journal articles on predicting storage behavior in energy-related journals in the Scopus database from 2010 to 2020, categorized by AI-based methodology.*

The field of prediction models for storage operation using machine learning is not yet established on a large scale though has made a significant jump in 2020. Neural networks are the most frequently applied method but also support vector machines, Gaussian process regression, and decision trees are used. In the field of neural networks, a trend towards RNNs is emerging, even if this is not yet clearly visible in the figures of the publications. Wang et al. [192] developed a prediction model for distributed electric heating storage systems. They find that their correlation-based LSTM model outperforms support vector machines and regular RNN models. Xiao et al. [193] come to a similar conclusion when comparing multiple methods for behavior learning in microgrids. They find LSTM models most suitable for microgrids that include storage systems.

[10] For details on the search term specification, see Table 15 in the appendix.

# 6 Field of application: Metamodeling and design of experiments for boosting the scenario scope

The basic experimental design method has been successfully used in combination with metamodeling in many publications. In the following, a possible approach is presented, as well as the benefits of metamodeling based on exemplary publications. Further, modeling recommendations for the combination of metamodeling and DOE are provided.

Unlike in real-world experiments, the input variables in computer-based experiments can be varied continuously with less effort. This allows the design space to be sampled at a higher resolution. However, in many use cases, simulations have a rather complex nature with non-linear behavior. Metamodeling using a full factorial experimental design (full-FD) and linear or polynomial regression is not possible in these cases. In these cases, more complex experimental designs (see section 4.2) and more complex methods for metamodeling, e.g. from the field of machine learning (s. section 3.1), must be used.

Figure 23 shows the results of the systematic literature review on the application of AI-based methods for metamodeling in energy-related publications.[11]

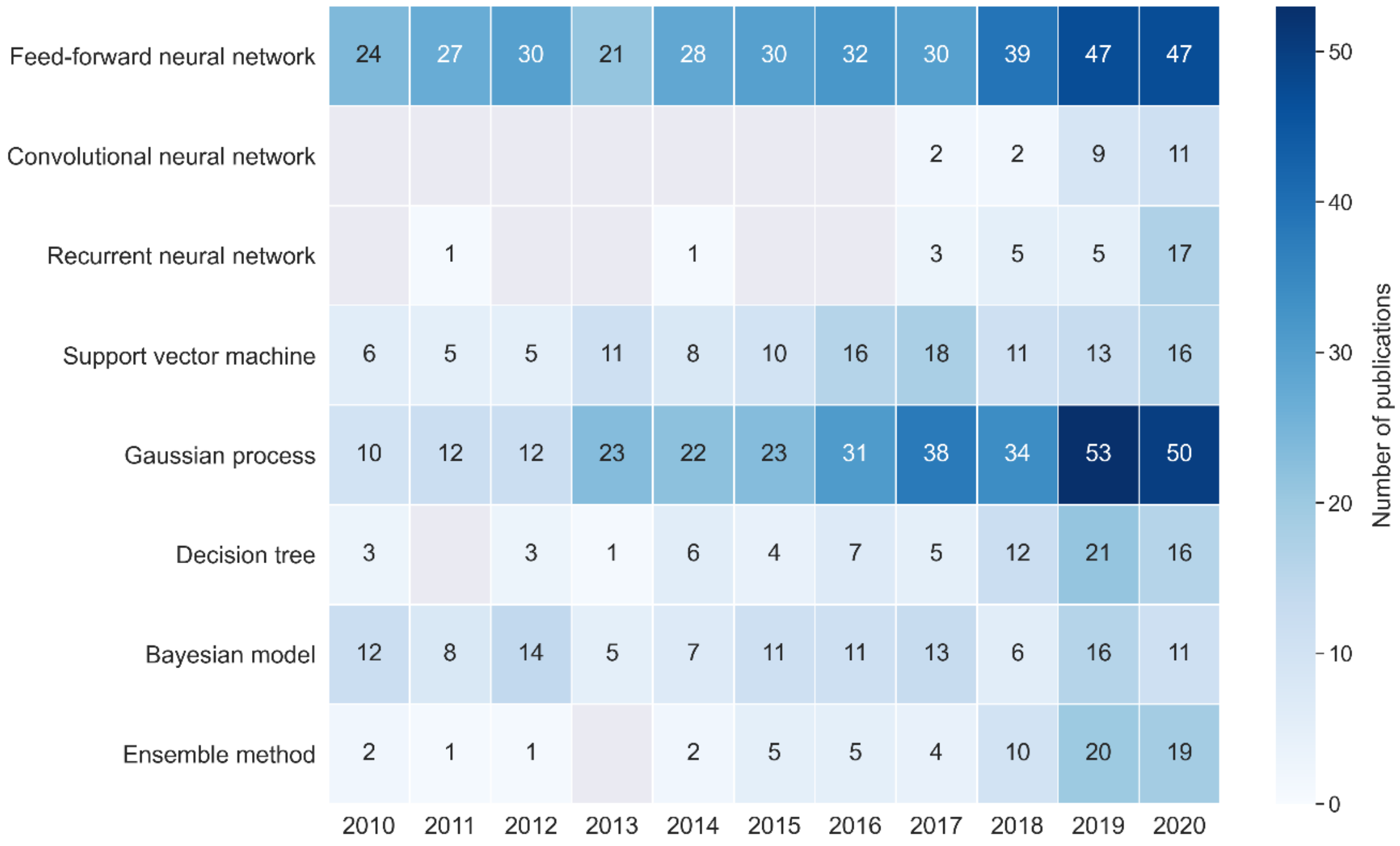

*Figure 23: Scientific journal articles on metamodeling in energy-related journals in the Scopus database from 2010 to 2020, categorized by metamodeling methodology.*

[11] For details on the search term specification, see Table 16 in the appendix.

In the past ten years, especially models based on FFNN and GPR have been used for metamodeling in publications in energy-related journals. Other methods of deep learning such as CNNs and RNNs have only been used from 2017 onwards. In addition to these methods, the number of publications per year applying SVMs and Bayesian models for metamodeling is relatively constant. A trend can be seen for the application of decision trees and ensemble methods that have become more popular since 2017.

Figure 24 shows the results of the systematic literature review on metamodeling applications in combination with DOE in energy-related publications.[12]

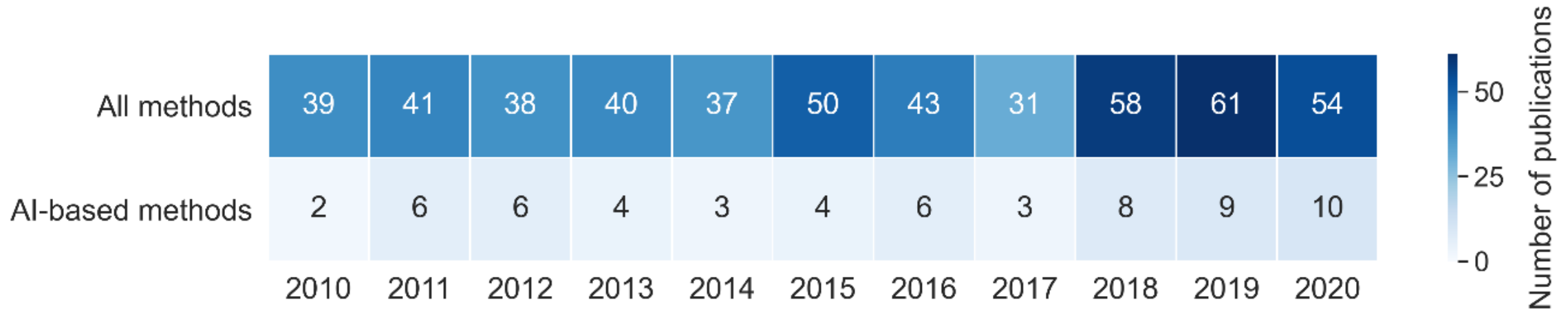

*Figure 24: Scientific journal articles on metamodeling in combination with design of experiments in energy-related journals in the Scopus database from 2010 to 2020, categorized by metamodeling methodology.*

Metamodeling and DOE are combined constantly since 2010 with numbers between ~35 and ~60 publications per year. AI-based methods only make up for a small share of these publications and the numbers are only slowly increasing. In 2020, ~20% of all publications we found on metamodeling and DOE in energy-related journals included the application of AI-based methods.

Storti et al. [194] apply a metamodeling approach for the optimization of the shape of ector plates for a drag-driven vertical axis Savonius wind turbine. Here, a two-dimensional computational fluid dynamics (CFD) model is metamodeled using an ANN and an LHD. To reduce the reverse moment of the turbine, the size and shape of the deflector plates are optimized. A genetic optimizer was used, which requires many results from a wide variety of scenarios. This can only be implemented with metamodeling, otherwise, the results have to be generated with many time-consuming simulations. Using an ANN-based approach, a regression coefficient $R^2$>0.97 on the training data and an $R^2$>0.95 on the validation data were obtained. Nolting et al. [11] show that linear regression and full factorial experimental designs should not be excluded in advance from computer simulations. In this comparative study, the prediction performance of linear regression (with full-FD) and artificial neural networks (with LHD) for the approximation of a probabilistic simulation model for assessing the security of electricity supply in Germany were investigated. The investigation showed that linear regression has better prediction quality for the present use case with fewer trial points and thus reduced associated simulation time. The results also show that, analogous to the selection of a suitable approximation method, a suitable method must be selected from the design of experiments on an application-specific basis.

A comprehensive comparison of metamodeling methods is presented by Østergård et al. [195]. In this study, the six most commonly used methods for metamodeling are applied to 13 application cases with different dimensionality and complexity. The authors conclude that the choice of method depends largely on the problem and that literature cannot provide a general preference for the method. From their results, the authors draw the following general conclusions, among others:

[12] For details on the search term specification, see Table 17 and Table 18 in the appendix.

- Standard settings generally provide poor or mediocre accuracies, so optimization of the hyperparameters is necessary,
- hyperparameters must be adapted to the respective problem,
- in general, the best results were achieved with GPR followed by ANNs, and multivariate adaptive regression splines (MARS),
- linear regression models achieved the worst accuracy due to the nonlinearity of the problems considered,
- for large datasets, ANNs performed most effectively, while GPR was slow and less robust,
- dimensionality has only a small influence on accuracy.

The authors impressively demonstrate the possible prediction quality of metamodeling methods. A coefficient of determination of up to $R^2$>0.99 was obtained for the eight mathematical benchmarks. For the building performance simulation problems, an $R^2$>0.90 for $CO_2$–emissions and $R^2$>0.99 for the remaining output parameters could be achieved.

# 7 Conclusion: Strategic benefits of AI and DoE for the assessment of security of electricity supply

Having demonstrated a broad variety of methods from the field of artificial intelligence that can be applied to energy system modeling in general and assessing security of electricity supply in particular, one core finding can be highlighted: As the necessity for and complexity of assessments of resource adequacy increase, combining AI-based methods with an adequate design of experiments offers the possibility for efficient metamodeling of complex energy system models. Hence, a broad variety of scenarios can be investigated and prevailing limits regarding runtime and hardware requirements can be efficiently circumvented while maintaining high degrees of accuracy.

In addition to that, we identified several potential fields of applications for the introduced AI-based methods within different steps in the model toolchain:

- Preprocessing of input data and data consolidation
- Forecasting of relevant input data such as electricity loads, feed-in from renewable energy sources, electricity prices, availabilities of power plants, and storage operations

While some of these fields have already received attention from the scientific community and there are many relevant publications, others have merely been investigated. In particular, forecasts regarding storage operation and the (non-)availability of individual power plants are rare, while there is a comprehensive body of literature on electricity load forecasts and the feed-in of renewables (see Figure 25).

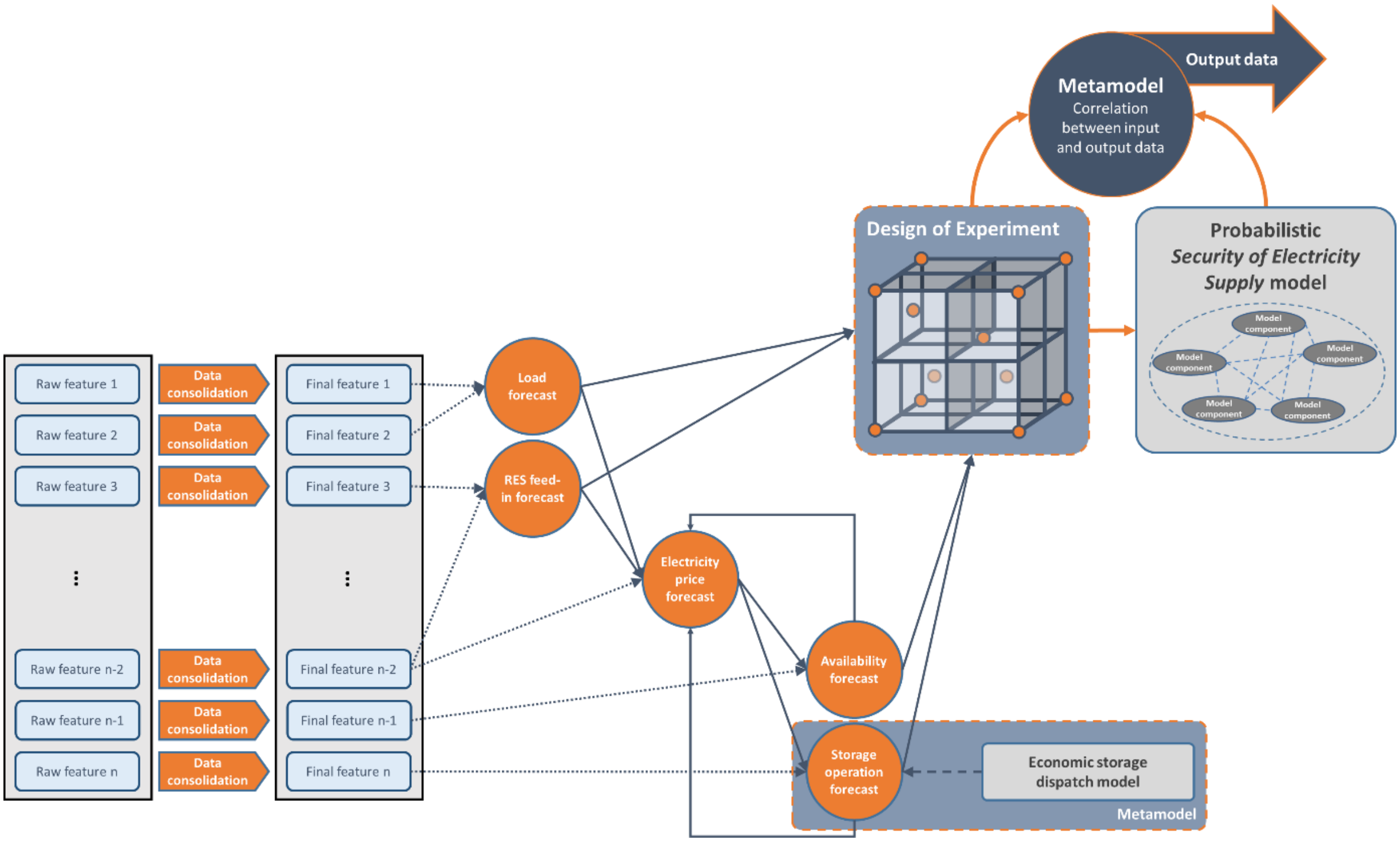

*Figure 25: Link of the different fields of application of machine learning and design of experiment methods in the context of the assessment of security of supply.*

Overall, we conclude that there is the necessity for future research regarding (1) the efficient metamodeling of complex models to assess security of electricity supply using AI-based methods and (2) applications of AI-based methods for forecasts of storage dispatch and (non-)availabilities as these are promising fields of application that have not sufficiently been covered, yet. Our review contributes by providing a quite comprehensive overview of candidate methods and potential fields of applications.

Regarding prevailing requirements for assessments of security of electricity supply [3], we find that the approach of AI-based metamodeling described in this review would be a beneficial supplement as it can help do depict the influences of prevailing uncertainties regarding the future development of necessary input data while allowing for a high level of detail of the model.

## Acknowledgements

This research was funded by the German Federal Ministry for Economic Affairs and Energy (BMWi) within the project KIVi (grant ID:0 3EI1022A).

## Appendix

*Table 10: Search term specification for the application of clustering in energy research*

| TITLE-ABS-KEY | | | | |
|---|---|---|---|---|
| **Application field (1)** | **Application field (2)** | **Method (Cluster)** | **SUBJAREA** | **PUBYEAR** |
| energy | - | k-means (k-means) | Ener | 2010 |
| | | k-medoids (k-medoids) | | 2011 |
| | | hierarchical clustering (hierarchical clustering) | | 2012 |
| | | | | 2013 |
| | | | | 2014 |
| | | | | 2015 |
| | | | | 2016 |
| | | | | 2017 |
| | | | | 2018 |
| | | | | 2019 |
| | | | | 2020 |

*Table 11: Search term specification for the application of dimensionality reduction in energy research*

| TITLE-ABS-KEY | | | | |
|---|---|---|---|---|
| **Application field (1)** | **Application field (2)** | **Method (Cluster)** | **SUBJAREA** | **PUBYEAR** |
| energy | - | principle component analysis (PCA) | Ener | 2010 |
| | | principal component regression (PCA) | | 2011 |
| | | partial least squares regression (PCA) | | 2012 |
| | | PCA (PCA) | | 2013 |
| | | discriminant analysis (discriminant analysis) | | 2014 |
| | | autoencoder (autoencoder) | | 2015 |
| | | t-distributed stochastic neighbor embedding (t-SNE) | | 2016 |
| | | t-SNE (t-SNE) | | 2017 |
| | | | | 2018 |
| | | | | 2019 |
| | | | | 2020 |

*Table 12: Search term specification for use case "Forecasting electricity load profiles"*

| TITLE-ABS-KEY | | | | |
|---|---|---|---|---|
| **Application field (1)** | **Application field (2)** | **Method (Cluster)** | **SUBJAREA** | **PUBYEAR** |
| load forecasting | - | artificial neural network (Feed-forward neural network) | Ener | 2010 |
| load prediction | | feed-forward neural network (Feed-forward neural network) | | 2011 |
| | | back propagation neural network (Feed-forward neural network) | | 2012 |
| | | multilayer perceptron (Feed-forward neural network) | | 2013 |
| | | ANN (Feed-forward neural network) | | 2014 |
| | | convolutional neural network (Convolutional neural network) | | 2015 |
| | | CNN (Convolutional neural network) | | 2016 |
| | | recurrent neural network (Recurrent neural network) | | 2017 |
| | | RNN (Recurrent neural network) | | 2018 |
| | | long short term memory (Recurrent neural network) | | 2019 |
| | | LSTM (Recurrent neural network) | | 2020 |
| | | gated recurrent unit (Recurrent neural network) | | |
| | | GRU (Recurrent neural network) | | |
| | | support vector machine (Support vector machine) | | |
| | | gaussian process (Gaussian process) | | |
| | | decision tree (Decision tree) | | |
| | | classification tree (Decision tree) | | |
| | | regression tree (Decision tree) | | |
| | | bayesian network (Bayesian model) | | |
| | | bayesian net (Bayesian model) | | |
| | | naive bayes (Bayesian model) | | |
| | | bayesian classification (Bayesian model) | | |
| | | bayesian regression (Bayesian model) | | |
| | | bayesian belief network (Bayesian model) | | |
| | | Bagging (Ensemble method) | | |
| | | decision tree (Ensemble method) | | |
| | | random forest (Ensemble method) | | |
| | | boosting (Ensemble method) | | |
| | | xgboost (Ensemble method) | | |
| | | catboost (Ensemble method) | | |
| | | LightGBM (Ensemble method) | | |

*Table 13: Search term specification for use case "Forecasting renewable feed-in profiles"*

| TITLE-ABS-KEY | | | | |
|---|---|---|---|---|
| **Application field (1)** | **Application field (2)** | **Method (Cluster)** | **SUBJAREA** | **PUBYEAR** |
| renewable feed-in | - | artificial neural network (Feed-forward neural network) | Ener | 2010 |
| wind feed-in | | feed-forward neural network (Feed-forward neural network) | | 2011 |
| solar feed-in | | back propagation neural network (Feed-forward neural network) | | 2012 |
| pv feed-in | | multilayer perceptron (Feed-forward neural network) | | 2013 |
| hydro feed-in | | ANN (Feed-forward neural network) | | 2014 |
| wind power forecast | | convolutional neural network (Convolutional neural network) | | 2015 |
| solar power forecast | | CNN (Convolutional neural network) | | 2016 |
| pv power forecast | | recurrent neural network (Recurrent neural network) | | 2017 |
| hydro power forecast | | RNN (Recurrent neural network) | | 2018 |
| wind power prediction | | long short term memory (Recurrent neural network) | | 2019 |
| solar power prediction | | LSTM (Recurrent neural network) | | 2020 |
| pv power prediction | | gated recurrent unit (Recurrent neural network) | | |
| hydro power prediction | | GRU (Recurrent neural network) | | |
| | | support vector machine (Support vector machine) | | |
| | | gaussian process (Gaussian process) | | |
| | | decision tree (Decision tree) | | |
| | | classification tree (Decision tree) | | |
| | | regression tree (Decision tree) | | |
| | | bayesian network (Bayesian model) | | |
| | | bayesian net (Bayesian model) | | |
| | | naive bayes (Bayesian model) | | |
| | | bayesian classification (Bayesian model) | | |
| | | bayesian regression (Bayesian model) | | |
| | | bayesian belief network (Bayesian model) | | |
| | | Bagging (Ensemble method) | | |
| | | decision tree (Ensemble method) | | |
| | | random forest (Ensemble method) | | |
| | | boosting (Ensemble method) | | |
| | | xgboost (Ensemble method) | | |
| | | catboost (Ensemble method) | | |
| | | LightGBM (Ensemble method) | | |

*Table 14: Search term specification for use case “(non-)availabilities”*

| TITLE-ABS-KEY | | | | |
|---|---|---|---|---|
| **Application field (1)** | **Application field (2)** | **Method (Cluster)** | **SUBJAREA** | **PUBYEAR** |
| power plant"W/2"reliability | - | artificial neural network (Feed-forward neural network) | Ener | 2010 |
| power plant"W/2"outage | | feed-forward neural network (Feed-forward neural network) | | 2011 |
| power plant"W/2"maintenance | | back propagation neural network (Feed-forward neural network) | | 2012 |

| | | |
|---|---|---|
| power plant"W/2"availability | multilayer perceptron (Feed-forward neural network) | 2013 |
| power plant"W/2" unavailability | ANN (Feed-forward neural network) | 2014 |
| power plant"W/2" non-availability | convolutional neural network (Convolutional neural network) | 2015 |
| | CNN (Convolutional neural network) | 2016 |
| | recurrent neural network (Recurrent neural network) | 2017 |
| | RNN (Recurrent neural network) | 2018 |
| | long short term memory (Recurrent neural network) | 2019 |
| | LSTM (Recurrent neural network) | 2020 |
| | gated recurrent unit (Recurrent neural network) | |
| | GRU (Recurrent neural network) | |
| | support vector machine (Support vector machine) | |
| | gaussian process (Gaussian process) | |
| | decision tree (Decision tree) | |
| | classification tree (Decision tree) | |
| | regression tree (Decision tree) | |
| | bayesian network (Bayesian model) | |
| | bayesian net (Bayesian model) | |
| | naive bayes (Bayesian model) | |
| | bayesian classification (Bayesian model) | |
| | bayesian regression (Bayesian model) | |
| | bayesian belief network (Bayesian model) | |
| | Bagging (Ensemble method) | |
| | decision tree (Ensemble method) | |
| | random forest (Ensemble method) | |
| | boosting (Ensemble method) | |
| | xgboost (Ensemble method) | |
| | catboost (Ensemble method) | |
| | LightGBM (Ensemble method) | |

*Table 15: Search term specification for use case “Forecasting storage operation”*

| TITLE-ABS-KEY | | | | |
|---|---|---|---|---|
| **Application field (1)** | **Application field (2)** | **Method (Cluster)** | **SUBJAREA** | **PUBYEAR** |
| storage"W/3"operation | pumped | artificial neural network (Feed-forward neural network) | Ener | 2010 |
| storage"W/3"dispatch | compressed air | feed-forward neural network (Feed-forward neural network) | | 2011 |
| storage"W/3"behavior | CAES | back propagation neural network (Feed-forward neural network) | | 2012 |
| storage"W/3"heuristic | battery | multilayer perceptron (Feed-forward neural network) | | 2013 |
| storage"W/3"schedule | BES | ANN (Feed-forward neural network) | | 2014 |
| | BESS | convolutional neural network (Convolutional neural network) | | 2015 |
| | flywheel | CNN (Convolutional neural network) | | 2016 |
| | heat | recurrent neural network (Recurrent neural network) | | 2017 |
| | thermal | RNN (Recurrent neural network) | | 2018 |
| | electricity | long short term memory (Recurrent neural network) | | 2019 |
| | energy | LSTM (Recurrent neural network) | | 2020 |
| | power plant | gated recurrent unit (Recurrent neural network) | | |
| | | GRU (Recurrent neural network) | | |
| | | support vector machine (Support vector machine) | | |
| | | gaussian process (Gaussian process) | | |
| | | decision tree (Decision tree) | | |
| | | classification tree (Decision tree) | | |
| | | regression tree (Decision tree) | | |
| | | bayesian network (Bayesian model) | | |
| | | bayesian net (Bayesian model) | | |
| | | naive bayes (Bayesian model) | | |
| | | bayesian classification (Bayesian model) | | |
| | | bayesian regression (Bayesian model) | | |
| | | bayesian belief network (Bayesian model) | | |
| | | Bagging (Ensemble method) | | |
| | | decision tree (Ensemble method) | | |
| | | random forest (Ensemble method) | | |
| | | boosting (Ensemble method) | | |
| | | xgboost (Ensemble method) | | |
| | | catboost (Ensemble method) | | |
| | | LightGBM (Ensemble method) | | |

*Table 16: Search term specification for the application of metamodeling in energy research*

| TITLE-ABS-KEY | | | | |
|---|---|---|---|---|
| **Application field (1)** | **Application field (2)** | **Method (Cluster)** | **SUBJAREA** | **PUBYEAR** |
| meta model | - | artificial neural network (Feed-forward neural network) | Ener | 2010 |
| metamodel | | feed-forward neural network (Feed-forward neural network) | | 2011 |
| metamodeling | | back propagation neural network (Feed-forward neural network) | | 2012 |
| metamodelling | | multilayer perceptron (Feed-forward neural network) | | 2013 |
| meta modeling | | ANN (Feed-forward neural network) | | 2014 |
| metamodelling | | convolutional neural network (Convolutional neural network) | | 2015 |
| | | CNN (Convolutional neural network) | | 2016 |
| | | recurrent neural network (Recurrent neural network) | | 2017 |
| | | RNN (Recurrent neural network) | | 2018 |
| | | long short term memory (Recurrent neural network) | | 2019 |
| | | LSTM (Recurrent neural network) | | 2020 |
| | | gated recurrent unit (Recurrent neural network) | | |
| | | GRU (Recurrent neural network) | | |
| | | support vector machine (Support vector machine) | | |
| | | gaussian process (Gaussian process) | | |
| | | decision tree (Decision tree) | | |
| | | classification tree (Decision tree) | | |
| | | regression tree (Decision tree) | | |
| | | bayesian network (Bayesian model) | | |
| | | bayesian net (Bayesian model) | | |
| | | naive bayes (Bayesian model) | | |
| | | bayesian classification (Bayesian model) | | |
| | | bayesian regression (Bayesian model) | | |
| | | bayesian belief network (Bayesian model) | | |
| | | Bagging (Ensemble method) | | |
| | | decision tree (Ensemble method) | | |
| | | random forest (Ensemble method) | | |
| | | boosting (Ensemble method) | | |
| | | xgboost (Ensemble method) | | |
| | | catboost (Ensemble method) | | |
| | | LightGBM (Ensemble method) | | |

*Table 17: Search term specification for the application of any type of metamodel in combination with design of experiments in energy research*

| TITLE-ABS-KEY | | | SUBJAREA | PUBYEAR |
|---|---|---|---|---|
| **Application field (1)** | **Application field (2)** | **Method (Cluster)** | | |
| meta model | design of experiment | - | Ener | 2010 |
| metamodel | | | | 2011 |
| metamodeling | | | | 2012 |
| metamodelling | | | | 2013 |
| meta modeling | | | | 2014 |
| metamodelling | | | | 2015 |
| | | | | 2016 |
| | | | | 2017 |
| | | | | 2018 |
| | | | | 2019 |
| | | | | 2020 |

*Table 18: Search term specification for the application of AI-based metamodels in combination with design of experiments in energy research*

| TITLE-ABS-KEY | | | SUBJAREA | PUBYEAR |
|---|---|---|---|---|
| **Application field (1)** | **Application field (2)** | **Method (Cluster)** | | |
| meta model | design of experiment | artificial neural network (Feed-forward neural network) | Ener | 2010 |
| metamodel | | feed-forward neural network (Feed-forward neural network) | | 2011 |
| metamodeling | | back propagation neural network (Feed-forward neural network) | | 2012 |

| | | |
|---|---|---|
| metamodelling | multilayer perceptron (Feed-forward neural network) | 2013 |
| meta modeling | ANN (Feed-forward neural network) | 2014 |
| metamodelling | convolutional neural network (Convolutional neural network) | 2015 |
| | CNN (Convolutional neural network) | 2016 |
| | recurrent neural network (Recurrent neural network) | 2017 |
| | RNN (Recurrent neural network) | 2018 |
| | long short term memory (Recurrent neural network) | 2019 |
| | LSTM (Recurrent neural network) | 2020 |
| | gated recurrent unit (Recurrent neural network) | |
| | GRU (Recurrent neural network) | |
| | support vector machine (Support vector machine) | |
| | gaussian process (Gaussian process) | |
| | decision tree (Decision tree) | |
| | classification tree (Decision tree) | |
| | regression tree (Decision tree) | |
| | bayesian network (Bayesian model) | |
| | bayesian net (Bayesian model) | |
| | naive bayes (Bayesian model) | |
| | bayesian classification (Bayesian model) | |
| | bayesian regression (Bayesian model) | |
| | bayesian belief network (Bayesian model) | |
| | Bagging (Ensemble method) | |
| | decision tree (Ensemble method) | |
| | random forest (Ensemble method) | |
| | boosting (Ensemble method) | |
| | xgboost (Ensemble method) | |
| | catboost (Ensemble method) | |
| | LightGBM (Ensemble method) | |